%% file: PWEMoE_Draft.tex
\documentclass[review]{fcs}
\usepackage{bm}
\usepackage{nicefrac}
\usepackage{microtype}
\usepackage{placeins}
\usepackage{stfloats}
\usepackage{cuted}
\usepackage{flafter}
\usepackage{algorithm}
\usepackage{algpseudocode}
\usepackage{epstopdf}

\graphicspath{{./}{figure/}}

\newcommand{\tablebodyfont}{\centering\normalsize\setlength{\tabcolsep}{4pt}}
\newcommand{\fitwidth}[1]{%
  \begingroup
  \sbox0{#1}%
  \ifdim\wd0>\linewidth
    \resizebox{\linewidth}{!}{\usebox0}%
  \else
    \makebox[\linewidth][c]{\usebox0}%
  \fi
  \endgroup
}

\theoremstyle{definition}
\newtheorem{remark}{Remark}

\input{math_commands.tex}

\newcommand{\usefcstemplate}{}

\title{Towards Efficient Pareto Set Approximation via Weight-Ensembling Mixture of Experts}
\shorttitle{Pareto Set Approximation via Weight-Ensembling Mixture of Experts}

\author[1]{Anke~TANG}
\author[2]{Li~SHEN}
\author[1]{Yong~LUO}
\author[3]{Shiwei~LIU}
\author[4]{Han~HU}
\author[1]{Bo~DU}
\author[5]{Dacheng~TAO}

\address[1]{School of Computer Science, Wuhan University, Wuhan 430072, China}
\address[2]{School of Computer Science and Engineering, Sun Yat-sen University, Shenzhen 518063, China}
\address[3]{University of Oxford, Oxford OX1 4BH, UK}
\address[4]{School of Computer Science, Beijing Institute of Technology, Beijing 100081, China}
\address[5]{Nanyang Technological University, Singapore 639798, Singapore}

\fcssetup{
  received = {month dd, yyyy},
  accepted = {month dd, yyyy},
}

\input{section/abstract.tex}

\begin{document}

\input{section/introduction.tex}
\input{section/related_work.tex}
\input{section/method.tex}
\input{section/experiments.tex}
\input{section/conclusion.tex}

\fontsize{8.5pt}{10pt}\selectfont

\section*{Appendixes}
\appendix
\input{section/appendix/appendix.tex}

\FloatBarrier
\bibliographystyle{fcs}
\bibliography{references}

\end{document}

%% file: math_commands.tex
\def\eqref#1{equation~\ref{#1}}

\def\1{\bm{1}}

\def\vb{{\bm{b}}}

\def\vl{{\bm{l}}}

\def\vr{{\bm{r}}}

\def\mD{{\bm{D}}}

\def\mW{{\bm{W}}}

\DeclareMathAlphabet{\mathsfit}{\encodingdefault}{\sfdefault}{m}{sl}
\SetMathAlphabet{\mathsfit}{bold}{\encodingdefault}{\sfdefault}{bx}{n}



%% file: section/abstract.tex
\newcommand{\pwemoeabstractbody}{%
  Solving multi-objective optimization problems for large deep neural networks is a challenging task due to the complexity of the loss landscape and the expensive computational cost of training and evaluating models.
  Efficient Pareto front approximation of large models enables multi-objective optimization for various tasks such as multi-task learning and trade-off analysis.
  Existing algorithms for learning Pareto set, including (1) evolutionary, hypernetworks, and hypervolume-maximization methods, are computationally expensive and have restricted scalability to large models;
  (2) Scalarization algorithms, where a separate model is trained for each objective ray, which is inefficient for learning the entire Pareto set and fails to capture the objective trade-offs effectively.
  Inspired by the recent success of model merging, we propose a practical and scalable approach to Pareto set learning problem via a weight-ensembling mixture of experts (MoE) structure.
  By ensembling the weights of specialized single-task models, the MoE module can effectively capture the trade-offs between multiple objectives and closely approximate the entire Pareto set of large neural networks.
  Once the routers are learned and a preference vector is set, the MoE module can be unloaded, thus no additional computational cost is introduced during inference.
  We conduct extensive experiments on vision and language tasks using large-scale models such as CLIP-ViT and GPT-2.
  The experimental results demonstrate that our method efficiently approximates the entire Pareto front of large models.
  Using only hundreds of trainable parameters of the MoE routers, our method even has lower memory usage compared to linear scalarization and algorithms that learn a single Pareto optimal solution, and are scalable to both the number of objectives and the size of the model.
  Our method significantly reduces the computational burden of learning the Pareto set, for example, in the two-task case, it can be achieved in just a few minutes.%
}

\ifdefined\usefcstemplate
\begin{abstract}
  \pwemoeabstractbody
\end{abstract}
\else
\abstract{\pwemoeabstractbody}
\fi

\keywords{Model Reuse, Model Merging, Mixture of Experts, Multi-Objective Optimization}

%% file: section/introduction.tex
\ifdefined\usefcstemplate
\else
\input{section/abstract.tex}
\maketitle
\fi

\section{Introduction}
\label{sec:introduction}

Multi-objective optimization problems (MOOPs) are ubiquitous in machine learning, where multiple objectives need to be optimized simultaneously.
The most related problem is multi-task learning, where a model learns a shared representation beneficial to all tasks by optimizing a joint objective function~\cite{ruderOverviewMultiTaskLearning2017,zhangSurveyMultiTaskLearning2021,jiang2026mitigating}. Typically, this function is an equally weighted sum of task-specific objectives.
However, conflicting objectives among tasks can lead to negative transfer, making finding a single model that excels in all tasks challenging or impractical.
In such scenarios, the Pareto set, containing all non-dominated solutions, is a better representation of the trade-offs between objectives~\cite{schafflerStochasticMethodSolution2002,burkeSearchMethodologiesIntroductory2014,senerMultiTaskLearningMultiObjective2019}.
A solution is said to be non-dominated if there is no other solution that is better in all objectives.

Existing Pareto set learning algorithms can be broadly divided into two primary categories depending on their ability to learn the entire Pareto set within a single run.
The first category is capable of handling the MOOPs directly and can learn the entire Pareto set in a single run, but often struggle with the complexity of modern deep neural networks, such as evolutionary algorithms~\cite{debFastElitistMultiobjective2002,zhangMOEAMultiobjectiveEvolutionary2007}, hypernetworks-based methods~\cite{navonLearningParetoFront2021}, and hypervolume maximization methods~\cite{beumeSMSEMOAMultiobjectiveSelection2007,zhangHypervolumeMaximizationGeometric2023}.
The second category is exemplified by scalarization algorithms, such as linear scalarization and exact Pareto optimal search~\cite{mahapatraMultiTaskLearningUser2020}.
These methods transform a MOOP into a series of single-objective problems, each corresponding to a different combination of objective weights. However, the necessity to train a distinct model for each preference ray can be prohibitively resource-intensive.

%
%
However, the efficient learning of the Pareto set for large deep neural networks remains an open challenge due to the complexity of the loss landscape, which is characterized by a vast, high-dimensional parameter space and the inherent non-convexity of the optimization problem.
This complexity makes it difficult to navigate towards a set of solutions that are non-dominated in all objectives.
Moreover, the computational cost of training and evaluating large models is prohibitively expensive.
Each iteration of training involves a substantial number of operations, including forward and backward passes through the network. The evaluation cost also scales up with the model size and the number of objectives.
These challenges directly impact the scalability of existing algorithms, making them fail to learn the entire Pareto set for large models efficiently.

In this study, we propose a novel approach to efficiently approximate the entire Pareto set problem across multiple downstream tasks.
Inspired by recent work on multi-task model merging~\cite{tangMergingMultiTaskModels2024}, we adapt the weight-ensembling Mixture of Experts (MoE) structure to approximate the Pareto set of large neural networks, which is practical and scalable to both the number of objectives and the size of the model.
Specifically, this approach can be broken down into three steps:
(1) \textbf{model mixing}: we first mix different task-specific models into a single up-scaled MoE model, which captures a range of knowledge from all tasks.
(2) \textbf{router fine-tuning}: we iteratively update the routers of the MoE module to learn the trade-offs between different objectives.
At each step, we sample a preference vector randomly from the standard simplex and fine-tune the routers using gradient information.
(3) \textbf{MoE unloading and inference}: once the routers are learned and a preference vector is set, the MoE modules can be unloaded, and no additional computational cost is introduced during inference.

The main contributions of this work are as follows:
\begin{itemize}
  \item We propose a novel approach to efficiently approximate the entire Pareto set of large neural networks using a Pareto Weight-Ensembling Mixture of Experts (PWEMoE) structure.
  \item We introduce two training strategies, one based on linear scalarization and another based on exact Pareto optimal search that can effectively fine-tune the routers of MoE modules.
  \item We explore different up-scaling strategies, including using only the MLP modules or incorporating the Attention blocks as well, for the weight-ensembling process.
  \item We conduct extensive experiments on both vision and language tasks using large-scale models such as CLIP-ViT and GPT-2 to demonstrate the effectiveness of our method.
\end{itemize}

%% file: section/related_work.tex
\section{Related Work}
\label{sec:related_work}

\textbf{Pareto optimal/set learning and multi-task learning.}
Here we provide an overview of the existing methods for learning the Pareto set or single Pareto optimal solutions.
Learning the Pareto set is a fundamental task in multi-objective optimization, where trade-offs between multiple objectives need to be considered.
Multi-task learning methods, although not follows user-specified preferences, are closely related to MOOP thus are also included here~\cite{wang2026diversity,yang2026fedpd}.
Evolutionary algorithms~\cite{debFastElitistMultiobjective2002,zhangMOEAMultiobjectiveEvolutionary2007}, hypervolume-maximization methods~\cite{beumeSMSEMOAMultiobjectiveSelection2007,zhangHypervolumeMaximizationGeometric2023}, and multi-objective Bayesian optimization~\cite{linParetoSetLearning} are popular choices for approximating the Pareto set.
(Sener O, et al. 2019) \cite{senerMultiTaskLearningMultiObjective2019} were among the first to explore Pareto optimality in the context of deep learning, employing the Multiple Gradient Descent Algorithm (MGDA)~\cite{desideriMultiplegradientDescentAlgorithm2012}.
Pareto Multi-Task Learning (PMTL)~\cite{linParetoMultiTaskLearning2019}, Exact Pareto Optimal (EPO)~\cite{mahapatraMultiTaskLearningUser2020}, and Impartial Multi-Task Learning (IMTL)~\cite{liuImpartialMultitaskLearning2020} enable users to adjust the trade-off between objectives to specify their preferences. However, they can only identify a single Pareto optimal solution in a single run.
Hypernetwork-based methods~\cite{navonLearningParetoFront2021,hoangImprovingParetoFront2023} and manifold learning-based methods~\cite{dimitriadisParetoManifoldLearning2023} can scale to deep neural networks with a few million parameters and learn the entire Pareto set simultaneously.
More recently, Pareto Low-Rank Adapters (PaLoRA)~\cite{dimitriadisParetoLowRankAdapters2025} reduce the memory overhead of Pareto front learning by attaching one low-rank adapter per objective to a shared backbone and combining them with the preference vector, together with a deterministic annealed preference schedule.
On the scalarization side, smooth Tchebycheff scalarization~\cite{linSmoothTchebycheffScalarization2024} and its set-based extension~\cite{linFewManyTchebycheff2025} provide gradient-friendly surrogates that, unlike linear scalarization, can also reach non-convex parts of the Pareto front.

\textbf{Model merging.}
Model merging is a highly effective and scalable approach to aggregating knowledge from multiple models by merging them into a unified model.
A common practice to merge multiple models is by performing element-wise interpolation on weights, such as simple averaging~\cite{wortsmanModelSoupsAveraging2022,chronopoulouAdapterSoupWeightAveraging2023,kaddourStopWastingMy2022,sanyalUnderstandingEffectivenessEarly2023}, task arithmetic~\cite{ilharcoEditingModelsTask2023,guillermoortiz-jimenezTaskArithmeticTangent2023}, and Fisher merging~\cite{matenaMergingModelsFisherWeighted2022}.
Other approaches such as mode connectivity-based methods~\cite{danielfreemanTopologyGeometryHalfrectified2017,draxlerEssentiallyNoBarriers2019,frankleLinearModeConnectivity2020,entezariRolePermutationInvariance2022,tatroOptimizingModeConnectivity2020,yunisConvexityLinearMode2022,bentonLossSurfaceSimplexes2021}, alignment-based methods~\cite{jinDatalessKnowledgeFusion2023,liuDeepNeuralNetwork2022,georgestoicaZipItMergingModels2023,ainsworthGitReBasinMerging2023} are also crucial for effective model merging.
A common challenge in model merging is task interference and parameter conflicts, which can be addressed through various approaches, such as subspace-based~\cite{yuLanguageModelsAre2023,tangConcreteSubspaceLearning2023}, representation-based~\cite{yangRepresentationSurgeryMultiTask2024,jinDatalessKnowledgeFusion2023}, and parameter-based methods~\cite{yadavResolvingInterferenceWhen2023}.
For a comprehensive review of model merging and its applications, please refer to~\cite{zhengLearnModelFineTuning2023,liDeepModelFusion2023,yadavSurveyModelMoErging2024,mao2025survey}.
Model upscaling methods~\cite{tangMergingMultiTaskModels2024,ostapenkoModularLLMsBuilding2024,tangSMILEZeroShotSparse2024} are also closely related to model merging, where a small model is upscaled to a larger model by leveraging the knowledge from multiple tasks.

\textbf{Preference-aware and controllable model merging} is the most related work to our work, yielding a continuum of trade-offs after one merge.
Rewarded soups~\cite{rameRewardedSoupsInterpolating2023} interpolate fine-tuned weights along the preference vector but degrade in the simplex interior.
MAP~\cite{liMAPLowcomputeModel2025} fits quadratic surrogates of merging coefficients yet must be re-fit per task subset, with cost growing in the number of objectives.
Pareto Merging~\cite{chenParetoMergingMultiObjective2025} and ReACT~\cite{wuParameterRepresentationClosedForm2026} attach preference-conditioned low-rank residuals or closed-form representation corrections on top of a fixed pre-merged backbone (e.g., AdaMerging).
In contrast, our method makes the merging itself preference-dependent by routing over per-task weight deltas.


Our proposed method differs from existing model merging approaches that usually integrate individual models into a single multi-task model, enabling efficient trade-off learning between multiple objectives.
Instead, we merge multiple expert models to approximate the entire Pareto set.
Compared with the concurrent preference-aware merging methods discussed above, the preference vector in our method does not act on a fixed merged backbone but drives a router that produces the merging coefficients of every up-scaled module, so the whole encoder is a function of the preference.
This makes the Pareto set approximation more expressive at a comparable parameter overhead, while remaining far cheaper than Pareto set learning methods that keep one full model per objective.

%% file: section/method.tex
\section{Reinterpreting Pareto Set Learning}

\begin{figure*}[t]
  \centering
  \includegraphics[width=0.85\linewidth]{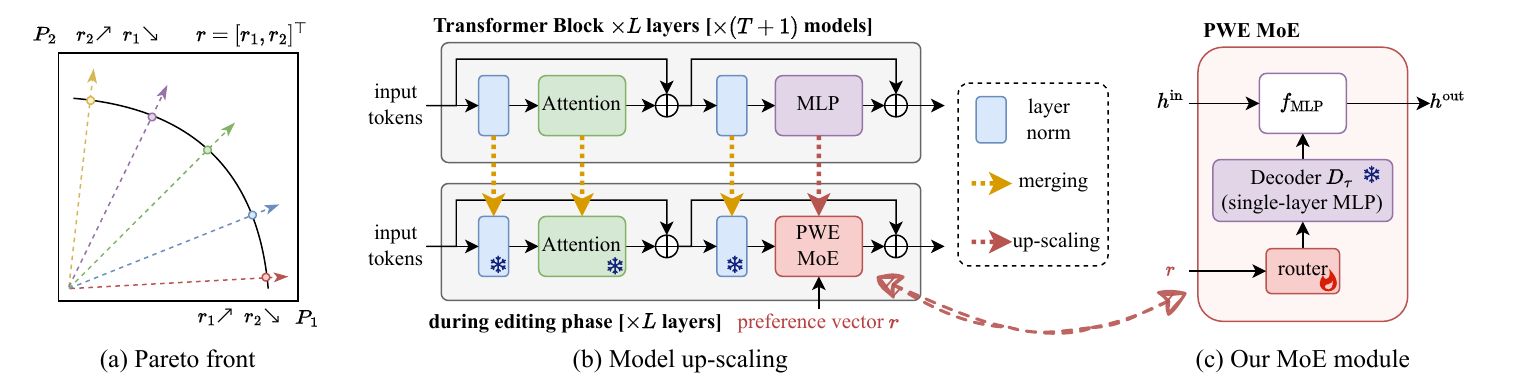}
  \caption{
    \textbf{Overview of the proposed method.}
    (a) An illustration of Pareto front learning in MOOP. Where $P_1$ and $P_2$ are performance metrics for two tasks, colored lines represent different Pareto optimal solutions, and the solid black line represents the Pareto front.
    (b) An overview of the model up-scaling process.
    We upcycle the MLP modules to MoE modules and merge the remaining parts using task arithmetic.
    (c) The MoE module, comprising a routing network and a parameter decoder network.
    The routing network accepts a user preference vector and generates routing weights for weight-ensembling.
  }
  \label{fig:framework}
\end{figure*}

In this section, we first introduce the formulation of Pareto set learning, then recast the Pareto set learning problem into the general framework of the model merging problem.

\subsection{Preliminary}
\label{sec:preliminary}

The challenge of multi-objective optimization problems (MOOP) stems from the conflicting objectives they present, resulting in a range of trade-off optimal solutions (referred to as Pareto optimal solutions) rather than a single optimal solution.
Given a set of $T$ objectives, each solution $\theta\in\mathbb{R}^{N}$ in the search space can be evaluated by a vector of $T$ objective functions $\vl(\theta) = [l_1(\theta), l_2(\theta), \ldots, l_T(\theta)]$. Before we delve into the details of our proposed method, we first introduce the concept of Pareto dominance, which is essential for comparing solutions in multi-objective optimization problems~\cite{boydConvexOptimization2004a,burkeSearchMethodologiesIntroductory2014}.
\begin{definition}[Pareto dominance]
  We say that a solution $\theta_1$ dominates another solution $\theta_2$, denoted as $\theta_1 \prec \theta_2$, if and only if $\forall t \in [T]$, $l_t(\theta_1) \leq l_t(\theta_2)$ and $\exists t \in [T]$ such that $l_t(\theta_1) < l_t(\theta_2)$.
\end{definition}
Pareto dominance is transitive, which means that if $\theta_1 \prec \theta_2$ and $\theta_2 \prec \theta_3$, then $\theta_1 \prec \theta_3$.
\begin{definition}[Pareto optimality, Pareto set, and Pareto front]
  A solution $\theta$ is said to be Pareto optimal if there is no other solution that dominates it. The set of all Pareto optimal solutions is called the Pareto set, and the corresponding objective vectors are called the Pareto front.
\end{definition}

\subsection{Pareto Set Learning as Multi-Task Model Merging Problem}

A popular solution to seek Pareto optimal solution is linear scalarization (LS), which converts the multi-objective optimization problem into a single-objective optimization problem by linearly combining the objectives, i.e., $\min_{\theta} \sum_{t=1}^{T} r_t l_t(\theta), \text{s.t.} \sum_{t=1}^T r_t=1 \text{ and } r_t > 0$, where $r_t$ is the weight for the $t$-th objective and the vector $\vr = [r_1, r_2, \ldots, r_T]$ is denoted as the preference vector.
We demonstrate the concept of the Pareto set/front using a two-task example, as shown in Figure~\ref{fig:framework}(a).
However, linear scalarization fails to capture the trade-offs between objectives and can only identify a single Pareto optimal solution along the convex part of the Pareto front~\cite{boydConvexOptimization2004a,navonLearningParetoFront2021}, and selecting an appropriate preference vector $\vr$ that accurately represents the user's preferences can be challenging.

Existing methods for learning the entire Pareto front face challenges in scaling to large models.
Pareto HyperNetwork is great and effective for small-scale models~\cite{navonLearningParetoFront2021}, but its practicality diminishes for larger models. This is because the output of the hypernetwork matches the number of parameters in the primary network. For instance, if the hypernetwork is a simple MLP with a single hidden layer of size $H=100$, its parameter count would be $O(H \times N)$. Consequently, a small model with 10M parameters would necessitate a hypernetwork with 1B parameters. A possible compromise is to generate only a subset of the parameters and share the rest across the tasks. However, this would lead to a suboptimal solution, and the design of the hypernetwork would be nontrivial.
Another intriguing approach is to learn a simplex Pareto subspace by learning a matrix of parameters $\Theta=[\theta_1, \theta_2, \dots, \theta_m]^\top \in \mathbb{R}^{m\times N}$ and a task weighting matrix $\mW\in \mathbb{R}^{m\times T}$ simultaneously~\cite{dimitriadisParetoManifoldLearning2023}.
However, the assumption of a linear mapping from parameter space to objective space may not be valid, resulting in an imbalanced distribution of Pareto optimal relative to the preference vector. Additionally, optimizing such a large parameter matrix $\Theta$ demands substantial memory and computational resources, resulting in poor scalability with respect to both the number of tasks and the size of the model.

\textbf{Why do we opt to merge models?}
Considering the above challenges and limitations in scalability, capturing the trade-offs, and flexibility in selecting the preference vector, we opt to learn Pareto set by leveraging the inherent knowledge from different task-specific models instead of learning from scratch.
Our key insights are:
(1) each task-specific model can be viewed as a Pareto optimal corresponding to a special preference vector, e.g. $\vr = [1,0,0,\dots,0]$ for the first task.
(2) The Pareto set can be approximated by interpolating between the task-specific models in the parameter space, which is more scalable and flexible than learning from scratch.
(3) Merging models allow us to capture the trade-offs between objectives, even when the merging weights are non-convex.
Formally, Given a set of models $\{\theta_i\}_{i=1}^T$, we can merge them into a single model $\theta_{\text{merged}} = \mathcal{A}(\{\theta_i\}_{i=1}^T, w)$, where $\mathcal{A}$ is a merging algorithm, and $w$ is the algorithmic parameters that control the merging process. To approximate the Pareto set, we can substitute the algorithmic parameters with a function of the preference vector, i.e., $w = R(\vr)$ where $R: \mathbb{R}^T\rightarrow\mathbb{R}^{|w|}$ can be an arbitrary nonlinear function.
The resulting merged model $\theta_{\text{merged}}$ is then a function of the preference vector $\vr$ as follows:
\begin{equation}
  \theta_{\text{merged}} = \mathcal{A}(\{\theta_i\}_{i=1}^T, R(\vr)).
\end{equation}

\section{Towards Efficient Pareto Set Approximation via Weight-Ensembling MoE}
\label{sec:method}

Inspired by recent advances in multi-task model merging~\cite{liDeepModelFusion2023,tangMergingMultiTaskModels2024}, we propose an MoE-based approach to approximate the entire Pareto front of large models via weight-ensembling. Our method is scalable to both the number of objectives and the size of the model, and significantly reduces the computational demand, consuming even less memory compared to linear scalarization.

\subsection{MoE-Based Model Up-Scaling and Pareto Set Approximation}
\label{sec:weight_ensembling_mixture_of_experts}

Given $T$ task-specific models $\{\theta_i\}_{i=1}^T$ that are fine-tuned from a common pre-trained model $\theta_0$, we upscale the model by incorporating our proposed MoE module, which handles objective trade-offs, as shown in Figure~\ref{fig:framework}(b).
To make the notation concise, we denote the part of parameters that are to be up-scaled as $\phi_0, \phi_1, \dots, \phi_T$, where $\phi_0$ and $\phi_i$ represent the parameters of the pre-trained model and the $i$-th task-specific model, respectively.
The remaining parts that do not up-scale are merged using task arithmetic, which is a straightforward, effective, and scalable element-wise operation. Similarly, we denote these parameters as $\psi_0, \psi_1, \dots, \psi_T$, the merged parameters are $\psi^* = \psi_0 + \lambda \sum_{t=1}^{T} \psi_t$. Where $\lambda$ is a scalar scaling coefficient selected on a validation set.
We also discuss this partial up-scaling strategy, along with another strategy that up-scales Attention blocks as well.

\textbf{The Pareto Weight-Ensembling Mixture of Experts (PWEMoE) module.}
Figure~\ref{fig:framework}(c) shows the proposed PWE MoE module, which includes a routing network $R: \mathbb{R}^T\rightarrow\mathbb{R}^T$, which is a nonlinear mapping, and a parameter decoder $\mathbb{R}^T\mapsto\mathbb{R}^{\vert\phi\vert}$, and a realization function $F: \mathbb{R}^{\vert\phi\vert}\mapsto \mathcal{F}$, mapping the parameters to function space of the up-scaled module.
The routing network takes a user preference vector $\vr$ as input and generates routing weights for weight-ensembling. These routing weights are then used by the parameter decoder network to generate the parameters for the realization function. The decoder network itself is a fixed linear layer. The MoE is described mathematically as:
\begin{equation}
  w        = R(\vr), \,
  \phi^*   = \mD_\tau w + \phi_0, \,
  h^{out}  = F(\phi^*)(h^{in}),
\end{equation}
where $\mD_\tau\in\mathbb{R}^{\vert\phi\vert\times T}$ is a dictionary matrix that contains the task vectors of the up-scaled module, $\mD_\tau = [\tau_1, \tau_2, \ldots, \tau_T]$ and $\tau_i = \phi_i - \phi_0$, $h^{in}$ and $h^{out}$ are the input and output sequence of tokens.
We denote the parameters of the routing network as $\theta_R$.
The major difference between the proposed PWE MoE module and the weight-ensembling MoE from \cite{tangMergingMultiTaskModels2024} is that our routing network $R$ takes the preference vector $\vr$ as input, which allows the MoE module to generate parameters based on user preference and can be unloaded. This is highlighted in Figures~\ref{fig:framework}(b) and \ref{fig:framework}(c).
In contrast, the original weight-ensembling MoE generates routing weights based on input data, aiming to adapt to the input data distribution.

\textbf{Approximation of the Pareto set.}
Given a MOOP with $T$ objectives, the Pareto front is defined as $\mathcal{P} = \{\mathcal{L}(\phi) \mid \phi \in \Phi \text{ and } \nexists \phi' \in \Phi \text{ s.t. } \phi' \prec \phi \}$, where $\mathcal{L}(\phi) = (l_1(\phi), \ldots, l_T(\phi))$ is the vector of task-specific losses, and $\Phi=\mathbb{R}^{\vert\phi\vert}$ is the space of parameters $\phi$ for the upscaled module.

The proposed PWE MoE module searches a subspace of the parameter space that is the span of the task vectors centered at pre-trained weights, rather than the entire parameter space.
Let $\Phi_{\text{MoE}} = \{\phi \mid \phi = \mD_\tau w + \phi_0, w \in \mathbb{R}^T\}$ be the subspace of $\Phi$ that can be represented by the proposed MoE model, we have $\Phi_{\text{MoE}} \subset \Phi$.
We can also define the Pareto front of the MoE model as a function of the preference vector $\vr$ as follows:
\begin{equation}
  \begin{aligned}
    \mathcal{P}_{\text{MoE}}
    &= \bigl\{
      \mathcal{L}'(\vr)
      \mid \\
    &\quad
      \mathcal{L}'(\vr)
      = \mathcal{L}(\mD_\tau R(\vr) + \phi_0), \\
    &\quad
      \vr \in \Delta^{T-1},\ 
      \nexists \vr'
      \text{ s.t. }
      \mathcal{L}'(\vr') \prec \mathcal{L}'(\vr)
    \bigr\}.
  \end{aligned}
\end{equation}
In other words, the Pareto front of the MoE model is the set of loss vectors achieved by the MoE model for all possible preference vectors $\vr$ in the $(T-1)$-dimensional simplex, such that no other preference vector achieves a loss vector that Pareto dominates it.
To further establish the connection between the Pareto front and the MoE Pareto front, we provide a theoretical analysis, which guarantees that the MoE module can approximate the Pareto front of the up-scaled model with a bounded error.

\begin{theorem}[Existence of an error bound, proved in Appendix~\ref{appendix:proof_of_error_bound}]
  \label{theorem:existence_of_error_bound}
  Similar to the true Pareto front $\mathcal{P}$, let $\mathcal{P}_{\text{MoE}} = \{\mathcal{L}(\phi) \mid \phi \in \Phi_{\text{MoE}} \text{ and } \nexists \phi' \in \Phi_{\text{MoE}} \text{ s.t. } \phi' \prec \phi \}$ be the Pareto front estimated by the proposed MoE module.
  Assume that the loss functions $\{l_t\}_{t=1}^T$ are continuous, then for any $\mathcal{L}(\phi^*) \in \mathcal{P}_{\text{MoE}}$, there exists an $\mathcal{L}(\phi) \in \mathcal{P}$ and an $\epsilon > 0$ such that
  $\|\mathcal{L}(\phi^*) - \mathcal{L}(\phi)\|_2 \leq \sqrt{T} \epsilon$.
\end{theorem}
Furthermore, if each loss function $l_t$ is Lipschitz continuous with constant $L_t$, i.e. $\|l_t(\phi) - l_t(\phi')\| \leq L_t \|\phi - \phi'\|_2$ for all $\phi, \phi' \in \Phi$, then we can take $L=\sup \{L_t\}_{t=1}^T$ and obtain
the following bound:
\begin{equation}
  \|\mathcal{L}(\phi^*) - \mathcal{L}(\phi)\|_2 \leq \sqrt{T} L \|\phi^* - \phi\|_2.
\end{equation}
However, determining the Lipschitz constants of the loss functions is non-trivial and depends on the specific model and task.
In practice, the value of $\epsilon$ is less important than the fact that it exists and can be made arbitrarily small by increasing the density of the MoE subspace $\Phi_{\text{MoE}}$ within the entire parameter space $\Phi$. This density can be controlled by factors such as the number of models to be up-scaled and the diversity of the task vectors.
In addition to the theoretical guarantee, we have:

\begin{remark}[Scalability]
  The number of trainable parameters in the MoE module scales linearly with the number of tasks and is independent of the width of the pre-trained model. This allows efficient scaling to a large number of tasks without significantly increasing memory usage.
\end{remark}
\begin{remark}[Computational efficiency]
  Because the routing weight vector $w$ depends only on the preference vector $\vr$, the MoE module can be unloaded by substituting the MoE module with $F(\phi^*)$ after setting a preference vector, and no additional computational cost is introduced during inference.
\end{remark}
\begin{figure}[t]
  \centering
  \includegraphics[width=0.62\linewidth]{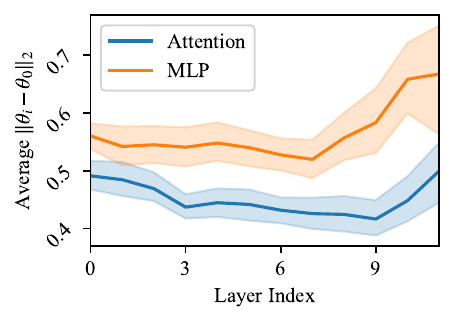}
  \caption{The $L_2$ norm of the delta parameters of CLIP.}
  \label{fig:l2_norm}
\end{figure}
\textbf{Two up-scaling strategies}.
We discuss two up-scaling strategies: one that only uses the MLP modules and another that incorporates the Attention blocks.
The choice of up-scaling strategy affects the quality of the Pareto set approximation and the computational demands.
A more fine-grained up-scaling strategy leads to better approximation but increases the number of trainable parameters in the routers.
To obtain a more comprehensive understanding of different strategies, we further investigate the similarity of Attention blocks and MLP modules in parameter space.

To establish a qualitative understanding of ``similarity'' in parameter space, we consider two functions are similar if they have similar outputs for similar inputs.
This definition can be extended to the parameter space of neural networks. To quantify this discrepancy, we Taylor expand the function $f(\cdot, \phi)$ around pretrained weights $\phi_0$:
\begin{equation}
  f(\cdot, \phi) = f(\cdot, \phi_0) + \nabla_\phi f(\cdot, \phi_0)^\top (\phi - \phi_0) + O(\|\phi - \phi_0\|^2).
\end{equation}
Where $\phi$ are the weights after fine-tuning.
For a given input $x$, the output of the function can be linearly approximated around $\phi_0$ since $\nabla_\phi f(x, \phi_0)$ is a constant vector. So $\|f(x, \phi_1) - f(x, \phi_2)\|_2 \propto \|\phi_1 - \phi_2\|_2$ for $\phi_1, \phi_2$ close to $\phi_0$. Therefore, the Euclidean distance in the parameter space is a good approximation of the functional discrepancy~\cite{tangMergingMultiTaskModels2024,lawsonMergingDecisionTransformers2023}.
Figure~\ref{fig:l2_norm} shows that Attention blocks are more similar to each other than MLPs, suggesting that up-scaling the MLPs may be more beneficial, as the MLPs are more diverse and thus task-specific.

\textbf{Structure design and parameter initialization of the routers.}
We implement the routing networks as fully connected feed-forward neural networks with a single hidden layer of size $2T$. The hidden layer is followed by a ReLU activation function. Mathematically, the routers are defined as
$R(\vr) = \mW_2 \mathop{ReLU}(\mW_1\vr + \vb_1) + \vb_2$,
where the weights $\mW_1$ and $\mW_2$ are initialized using the Gaussian distribution with zero mean and standard deviation of $0.01$, and the bias $\vb_1$ is initialized to zero, and $\vb_2$ is initialized to $\lambda$, where $\lambda$ is a same scalar scaling coefficient used in the task arithmetic.
After this initialization setup, routers are encouraged to output an initial routing weight vector that is close to the task arithmetic.

\begin{algorithm}[hbt]
  \small
  \caption{Router fine-tuning}\label{alg:router_training}
  \begin{algorithmic}[1]
    \While{not converged}
    \State $\vr \sim \mathop{Dir}(\alpha = 1)$
    \For {$m$ in all MoE modules}
    \State $\phi^* \gets \mD_\tau R(\vr) + \phi_0$
    \State unload $m$ to $F(\phi^*)$
    \EndFor
    \For {each task $t$}
    \State sample a batch of data $B_t \sim\mathcal{D}_t$
    \State compute the loss $l_t$ on $B_t$
    \EndFor
    \If {LS}
    \State $L \gets \sum_{t=1}^{T} r_t l_t$
    \ElsIf {EPO}
    \State $L \gets EPO(\theta_R, \{l_t\}_{t=1}^T, \vr)$
    \EndIf
    \State $\theta_R \gets \theta_R - \eta \nabla_{\theta_R} L$
    \State reset MoE modules
    \EndWhile
  \end{algorithmic}
\end{algorithm}


\subsection{Model Training}
\label{sec:model_training}

Once the model is merged and up-scaled, the routers of MoE modules undergo fine-tuning as described in Algorithm~\ref{alg:router_training}.
During this fine-tuning process, we uniformly sample $\vr$ from the $(T-1)$-simplex at each iteration, equivalent to sampling from a Dirichlet distribution with a concentration parameter $\alpha$, which is set to $1$ in our experiments.
We then merge the MoE models with the given preference vector. The loss function $L$ is computed in the same manner as linear scalarization (LS) or the expected Pareto optimal (EPO). Then the routers are updated using the gradient descent method.
This process is repeated until the routers converge.

The number of trainable parameters depends on the number of tasks and the extent to which the model is up-scaled, typically ranging from hundreds to thousands. Compared to full fine-tuning, the number of parameters is quite low. This results in our method having relatively low memory usage, even lower than linear scalarization.
In Table~\ref{table:training_details_image_classification}, we summarize the device states during training.

\subsection{Memory-Efficient Router Training}
\label{sec:memory_efficient}

\begin{figure*}[t]
  \centering
  \includegraphics[width=0.65\linewidth]{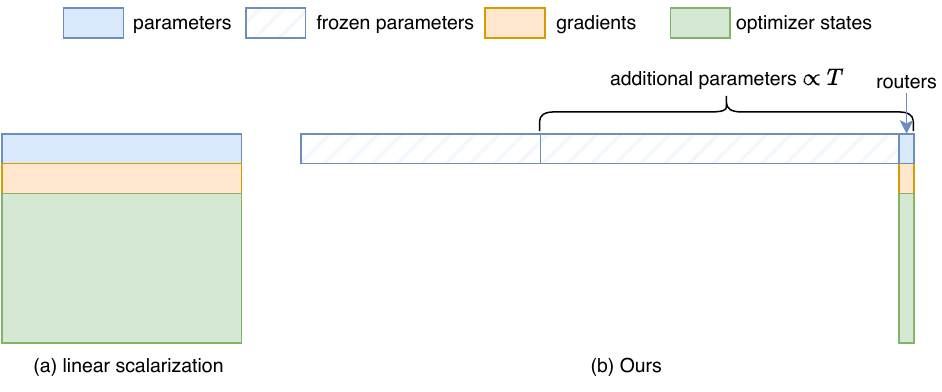}
  \caption{
    Comparison of memory consumption between linear scalarization and our proposed method.
  }
  \label{fig:memory_analysis}
\end{figure*}

In this section, we analyze the memory consumption of the proposed method and discuss possible strategies to further reduce memory usage.
We compare the memory consumption of the proposed method with linear scalarization, which is the most widely used method for multi-task learning, as shown in Figure~\ref{fig:memory_analysis}.
The additional parameters during training introduced by our method are the router $R$ (trainable) and the weights of the Docoder $\mD_\tau$ (fixed) in the PWE MoE module.
The router is a compact MLP with a non-linear activation function, take the 8-task experiment in Section~\ref{section:pareto_set_approximation_image_classification} as an example, the router has only 18.20K parameters, while the total parameter count of the upscaled MoE model is 1.02B.
As shown in Tables~\ref{table:training_details_image_classification_simplified} and \ref{table:training_details_image_classification}, our approach to approximating the Pareto set through model merging is highly memory-efficient, even when compared to the linear scalarization method, which introduces no additional parameters and learns only a single solution at a time.
This is because the memory consumption of model training encompasses model parameters, gradients and optimization states. For a model that is fully trainable, the optimization states are the dominant part of the memory consumption, which is proportional to the number of trainable parameters. Because for the upscaled MoE model, the router $R$ is only a small part of the model, the memory consumption of the gradients and optimization states are negligible. We illustrate this point in Figure~\ref{fig:memory_analysis}.
In fact, it is easy to further reduce the number of parameters of the MoE models using SVD to compress the weights of the Decoder $\mD_\tau$ in the PWE MoE module or combine with PEFT methods such as LoRA, but that is not the key insight of this work, so we leave it for future work.

%% file: section/experiments.tex
\section{Experiments}
\label{sec:experiments}

In this section, we conduct experiments on a variety of tasks, including image classification and text generation to demonstrate the effectiveness and scalability of our method.
Our experiments are conducted on two classes of large SOTA models: CLIP-ViT~\cite{radfordLearningTransferableVisual2021} for image classification tasks and GPT-2~\cite{radfordLanguageModelsAre2019} for text generation tasks, with up to eight objectives.
In terms of computational resources, our method requires only a few minutes to estimate the entire Pareto set.
Code for reproduction is available at \url{https://github.com/tanganke/pareto_set_learning}.

\textbf{Baseline methods.} We compare our method with the following baseline Pareto set learning methods: Single Task Learning (STL), LS, EPO~\cite{mahapatraMultiTaskLearningUser2020}, and MGDA~\cite{senerMultiTaskLearningMultiObjective2019}.
In addition, we compare our method with several multi-task model merging methods, including Weight Averaging~\cite{wortsmanModelSoupsAveraging2022}, Fisher Merging~\cite{matenaMergingModelsFisherWeighted2022}, RegMean~\cite{jinDatalessKnowledgeFusion2023}, Task Arithmetic~\cite{ilharcoEditingModelsTask2023}, and Ties-Merging~\cite{yadavResolvingInterferenceWhen2023} and AdaMerging~\cite{yangAdaMergingAdaptiveModel2023}, which produce a single ``one-size-fits-all'' model without preference control.
We also compare our method with Pareto Merging~\cite{chenParetoMergingMultiObjective2025}, which is a recent preference-aware model merging method that produces a Pareto set of models.

\subsection{Open-Vocabulary Image Classification}
\label{section:pareto_set_approximation_image_classification}

We first evaluate our method on open-vocabulary image classification tasks using the CLIP models, which are pre-trained on large-scale image-text pairs.
We consider eight downstream tasks spanning various data domains, including SUN397~\cite{xiao_sun_2010}, Stanford Cars~\cite{krause_3d_2013}, RESISC45~\cite{cheng_remote_2017}, EuroSAT~\cite{helber2018introducing}, SVHN~\cite{netzer_reading_2021}, GTSRB~\cite{stallkamp_man_2012}, MNIST~\cite{lecunGradientbasedLearningApplied1998} and DTD~\cite{cimpoi_describing_2014}.
A summarization of computational requirements is provided in Tables~\ref{table:training_details_image_classification_simplified} and \ref{table:training_details_image_classification}, demonstrating the computational efficiency of our method.

\begin{figure*}[t]
  \centering
  \includegraphics[width=0.90\linewidth]{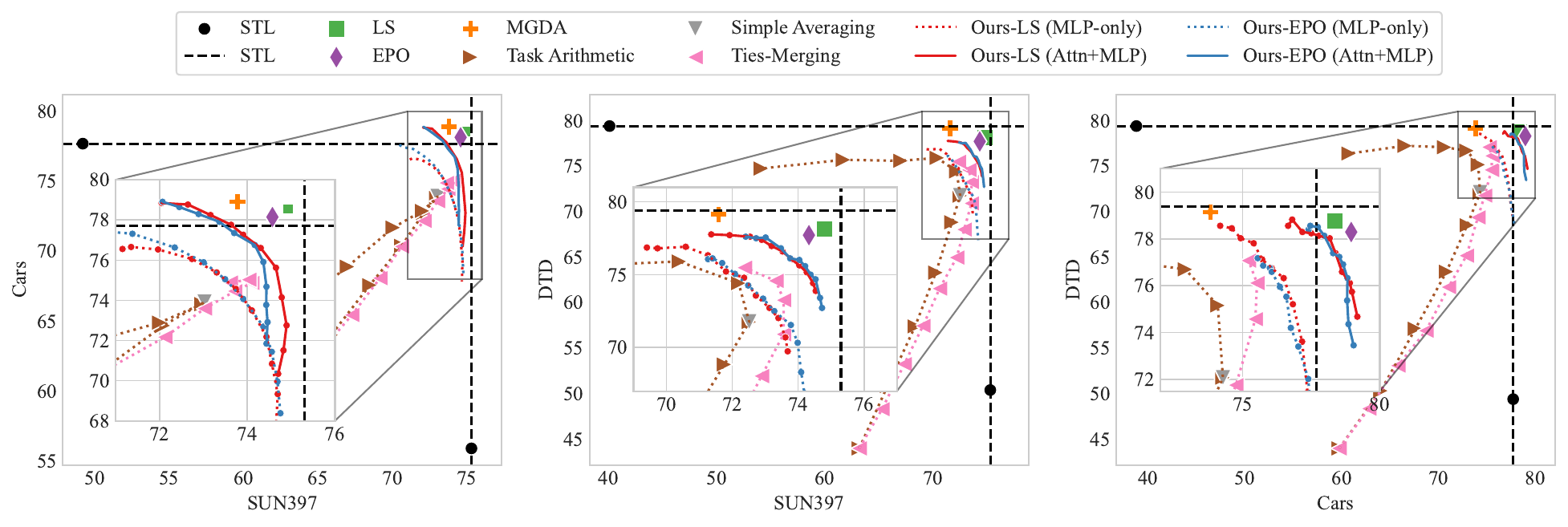}\\[0.5em]
  \includegraphics[width=0.90\linewidth]{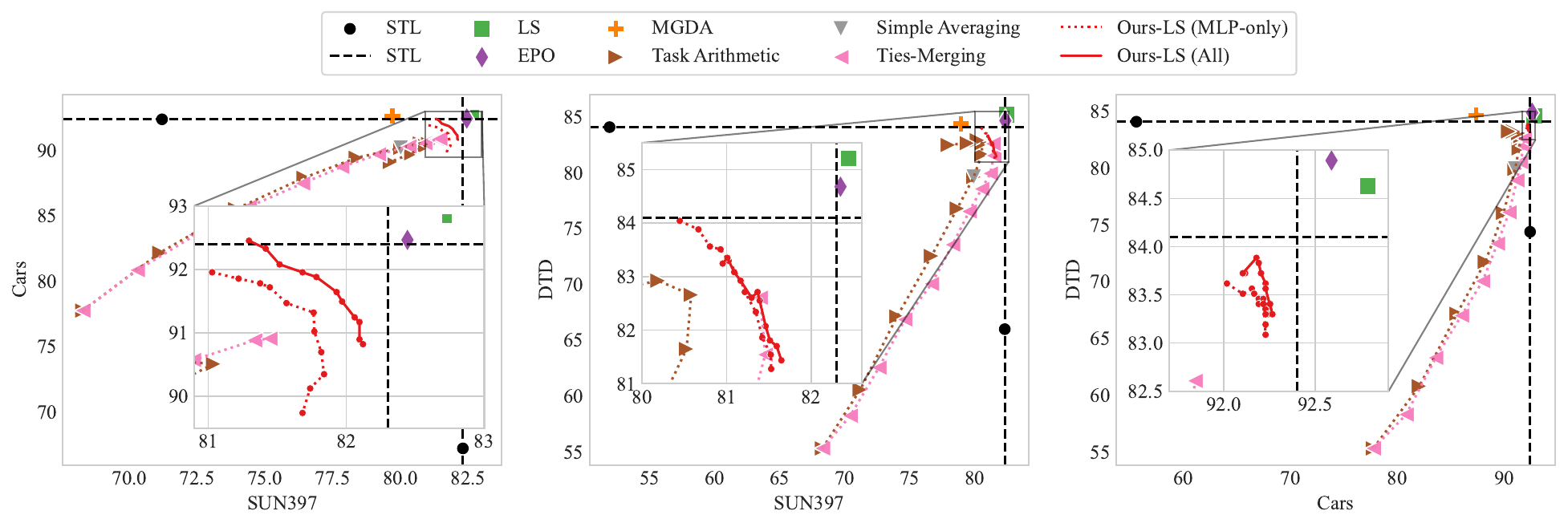}
  \caption{
    Visualization of two-task image classification results using CLIP-ViT-B/32 (top) and CLIP-ViT-L/14 (bottom).
  }
  \label{fig:clip_vitb32_pareto_moe_two_task}
\end{figure*}

\begin{figure}[hbt]
  \centering
  \includegraphics[width=0.6\linewidth]{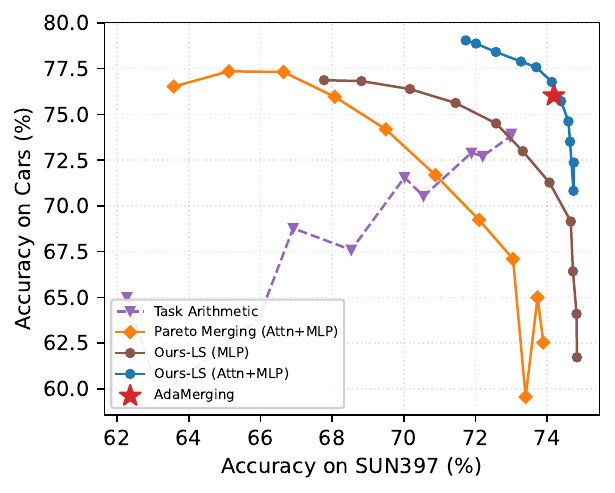}
  \caption{
    Pareto fronts on the SUN397-Cars pair with CLIP-ViT-B/32.
  }
  \label{fig:pareto_front_comparison}
\end{figure}

\textbf{Pairs of tasks.} Initially, we focus on Pareto set learning for two tasks, selecting the SUN397-Cars, SUN397-DTD, and Cars-DTD pairs for assessment.
We visualize the results using CLIP-ViT-B/32 and CLIP-ViT-L/14 in Figure~\ref{fig:clip_vitb32_pareto_moe_two_task}. The horizontal and vertical axes represent the accuracy of the first and second tasks, respectively. For Task Arithmetic and Ties-Merging, we show the results with different scaling factor.
In Table~\ref{table:hypervoulume} and Figure~\ref{fig:pareto_front_comparison}, we report the hypervolume and the shape of the approximated front on the SUN397-Cars pair, together with the recent preference-aware merging methods and AdaMerging.

\begin{table}[b]
  \centering
  \tablebodyfont
  \caption{
    Computational summary for two-task Pareto set learning using CLIP-ViT-B/32.
    Full details are in Table~\ref{table:training_details_image_classification}.
  }
  \label{table:training_details_image_classification_simplified}
  \begin{tabular}{@{}lcc@{}}
    \toprule
    Method              & GPU usage & Wall Time          \\
    \midrule
    LS (single point)   & 3.4GB     & $\approx$ 4 mins   \\
    EPO  (single point) & 6.1GB     & $\approx$ 8 mins   \\
    Ours-LS             & 2.8GB     & $\approx$ 2-3 mins \\
    Ours-LS (All)       & 3.9GB     & $\approx$ 3 mins   \\
    Ours-EPO            & 3.1GB     & $\approx$ 3-4 mins \\
    Ours-EPO (All)      & 3.9GB     & $\approx$ 5-6 mins \\
    \bottomrule
  \end{tabular}
\end{table}

\begin{table}[ht]
  \centering
  \tablebodyfont
  \caption{
    Hypervolume comparison on SUN397 and Cars using CLIP-ViT-B/32.
  }
  \label{table:hypervoulume}
  \begin{tabular}{@{}lc@{}}
    \toprule
    Method                    & Hypervolume     \\
    \midrule
    Pareto Merging (MLP)      & 0.5701          \\
    Pareto Merging (Attn+MLP) & 0.5671          \\
    \midrule
    Ours-LS (MLP)             & 0.5620          \\
    Ours-EPO (MLP)            & 0.5688          \\
    Ours-LS (Attn+MLP)        & \textbf{0.5999} \\
    Ours-EPO (Attn+MLP)       & 0.5872          \\
    \bottomrule
  \end{tabular}
\end{table}

We have the following key observations:
(1) For CLIP-ViT-B/32 models, joint fine-tuning on SUN397-Cars and Cars-DTD improves performance on Cars but worsens it on SUN397 and DTD, suggesting a one-way positive transfer to Cars. Joint tuning on SUN397 and DTD negatively impacts both, showing a negative transfer effect. \textit{Negative transfer is a common issue in multi-task learning}.
(2) For CLIP-ViT-L/14 models, performance improvements are generally positive for LS and EPO, highlighting that \textit{the larger model size enhances robustness to task interference.}
(3) Our method is able to successfully approximate the Pareto set, \textit{providing an good trade-off between the two tasks}. Ours-LS shows similar performance to Ours-EPO.
(4) For CLIP-ViT-B/32, up-scaling both the MLP and Attention leads to consistently better performance. However, for CLIP-ViT-L/14, gains from scaling both components are similar, showing that \textit{the benefits of up-scaling Attention are less significant in larger models.}
(5) Non-dominant points on the Task Arithmetic curves mainly appear in the middle, whereas the Ties-Merging curves are near the endpoints. This indicates that \textit{Ties-Merging is more effective at resolving task interference compared to Task Arithmetic}, aligning with the observations in Figure.~\ref{fig:clip_two_tasks_task_arithmetic_and_ties_merging}, where Ties-Merging consistently increases the average accuracy.

\begin{figure}[hbt]
  \centering
  \includegraphics[width=0.6\linewidth]{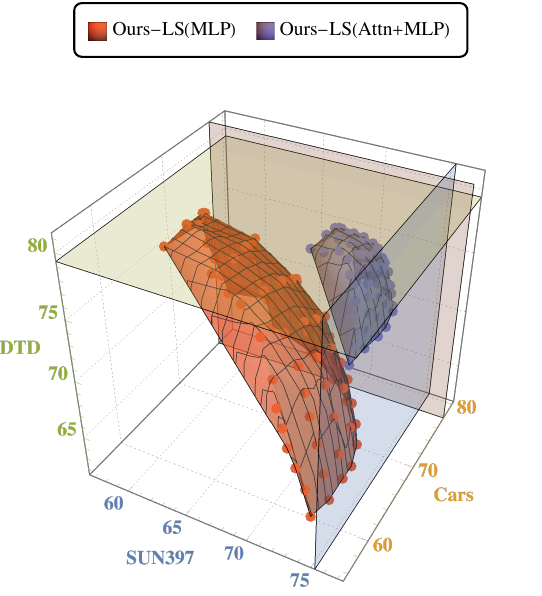}
  \caption{
    The 3D visualization of the Pareto front using CLIP-ViT-B/32 models.
    The planes represent the fine-tuned models.
  }
  \label{fig:ours_ls_3d}
\end{figure}

\textbf{Three tasks.} We further extend our experiments to three tasks using the SUN397-Cars-DTD triplet.
In Figure~\ref{fig:ours_ls_3d}, we visualize the 3D view of Pareto front for Ours-LS (MLP) and Ours-LS (Attn+MLP) using CLIP-ViT-B/32.
We uniformly sample preference vectors from the standard 2-dimensional simplex, generate parameters for the primary model and evaluate on the three tasks.
It is observed that the Pareto front is convex, and the uniformly sampled points are distributed nearly uniformly across the front.
The points on Ours-LS (Attn+MLP) generally dominate those on Ours-LS (MLP), indicating that \textit{up-scaling both components can improve the approximation of the Pareto front}.
Table~\ref{table:our_ls_3d} is a performance comparison table of different methods, utilizing an equal-weight preference vector and three special unit preference vectors.
Here are some key observations from the table:
(1) Fine-tuned (STL) models achieve the highest average accuracy of 77.5\% across the three tasks. They are served as the upper bounds of the single task accuracies here.
(2) Among model merging methods, Ours-LS(Attn+MLP) outperforms other model merging methods with the highest average accuracy of 74.2\%, followed by Ours-LS(MLP) at 70.3\%.
(3) Among model merging methods, when using Ours-LS (Attn+MLP) with unit preference vectors, the model achieves the highest accuracies on individual tasks. This demonstrates that \textit{by adjusting the preference vector, it can control the trade-off between objectives}.

\begin{table*}[t]
  \tablebodyfont
  \centering
  \caption{
    Three-task experiments using CLIP-ViT-B/32.
    In particular for Ours-LS (Attn+MLP), we report the performance of equal weights, three special unit preference vectors and three preference vectors that are not one-hot to demonstrate the learned trade-offs.
    The largest value and second largest value in each column are highlighted in \textbf{bold} and \textit{\underline{italic}}, respectively.}
  \label{table:our_ls_3d}
  \centering
  \begin{tabular*}{\linewidth}{@{\extracolsep{\fill}}lcccc}
    \toprule
    {Method}                                                & {SUN397}                  & {Cars}                    & {DTD}                     & {Avg.}                    \\
    \midrule
    Pretrained                                              & 63.2                      & 59.2                      & 44.4                      & 55.6                      \\
    Fine-tuned (STL)                                        & 75.3                      & 77.7                      & 79.4                      & 77.5                      \\
    \midrule
    LS (equal-weight)                                       & 75.1                      & 77.2                      & 78.9                      & 77.1                      \\
    EPO (equal-weight)                                      & 74.3                      & 78.6                      & 77.7                      & 76.9                      \\
    MGDA                                                    & 68.9                      & 65.7                      & 79.5                      & 71.4                      \\
    \midrule
    Weight Averaging                                        & 70.1                      & 69.7                      & 64.9                      & 68.2                      \\
    Task Arithmetic                                         & 70.1                      & 70.6                      & 67.6                      & 69.4                      \\
    Ties-Merging                                            & 71.3                      & 71.2                      & 67.7                      & 70.1                      \\
    Ours-LS (MLP,EW)                                        & 69.4                      & 68.8                      & 72.6                      & 70.3                      \\
    \textbf{Ours-LS (Attn+MLP,EW)}                          & 71.7                      & 74.7                      & 76.1                      & \textbf{74.2}             \\
    \textbf{Ours-LS (Attn+MLP,$[\mathbf{1},0,0]$)}          & \textbf{73.3}             & 65.7                      & 71.7                      & 70.2                      \\
    \textbf{Ours-LS (Attn+MLP,$[0,\mathbf{1},0]$)}          & 69.1                      & \textbf{77.9}             & 71.3                      & 72.8                      \\
    \textbf{Ours-LS (Attn+MLP,$[0,0,\mathbf{1}]$)}          & 68.2                      & 70.4                      & \textbf{77.1}             & 71.9                      \\
    \textbf{Ours-LS (Attn+MLP,$[\underline{0.6},0.2,0.2]$)} & \textit{\underline{73.2}} & 72.0                      & 74.0                      & 73.1                      \\
    \textbf{Ours-LS (Attn+MLP,$[0.2,\underline{0.6},0.2]$)} & 71.1                      & \textit{\underline{77.1}} & 74.1                      & \textit{\underline{74.1}} \\
    \textbf{Ours-LS (Attn+MLP,$[0.2,0.2,\underline{0.6}]$)} & 70.7                      & 73.8                      & \textit{\underline{76.9}} & 73.8                      \\
    \bottomrule
  \end{tabular*}
\end{table*}

\textbf{Scale up to eight tasks.} We also broaden our experiments to include all eight downstream tasks.
Due to the difficulty of visualizing the results in higher dimensions, we present the performance comparison in Table~\ref{table:clip_vitb32_eight_task} for CLIP-ViT-B/32 and CLIP-ViT-L/14.
For fine-tuned models, we report the task performance for each individual task, whereas for other methods, we report the performance using the same backbone.
In particular, for our method, we choose an equal-weight preference vector of $[\frac{1}{8},\dots,\frac{1}{8}]^\top$ to assess the multi-task performance.
The results of EPO and MGDA for CLIP-ViT-L/14 are not presented due to unsuccessful attempts to fine-tune the model with no more than 4$\times$4090 GPUs due to insufficient GPU memory with batch size set to 12 per task.
Due to the difficulty of visualizing the results in such a high-dimensional space, we present the performance comparison in Table~\ref{table:8tasks} using different preference vectors, including equal weights and eight one-hot vectors.

We observe that (1) model merging methods generally have lower average accuracies compared to multi-task learning methods.
(2) Moreover, the proposed methods, Ours-LS (MLP), Ours-LS (Attn+MLP), Ours-EPO (MLP), and Ours-EPO (Attn+MLP), outperform model merging baseline methods on both CLIP-ViT-B/32 and CLIP-ViT-L/14. (3) As the model size grows, the performance gap between multi-task learning methods and model merging methods narrows, indicating that \textit{task arithmetic is an emerging property of large models}. This is consistent with the conclusion in~\cite{guillermoortiz-jimenezTaskArithmeticTangent2023}. (4) Only up-scaling the MLP modules outperforms the other model merging methods, up-scaling both the Attention and MLP modules further improves the performance.

\begin{table*}[t]
  \centering
  \tablebodyfont
  \caption{
    Multi-task performance comparison on eight image classification tasks using CLIP-ViT-B/32 (top) and CLIP-ViT-L/14 (bottom).
  }
  \label{table:clip_vitb32_eight_task}
  \label{table:clip_vitl14_eight_task}
  \fitwidth{%
    \begin{tabular}{@{}lccccccccc@{}}
      \toprule
      {Method}            & {SUN397}      & {Cars}        & {RESISC45}    & {EuroSAT}     & {SVHN}        & {GTSRB}       & {MNIST}       & {DTD}         & {Avg.}        \\
      \midrule
      Pretrained          & 63.2          & 59.2          & 60.2          & 45.0          & 31.6          & 32.6          & 48.3          & 44.4          & 48.1          \\
      Fine-tuned (STL)    & 75.3          & 77.7          & 96.1          & 99.9          & 97.5          & 98.7          & 99.7          & 79.4          & 90.5          \\
      \midrule
      LS (equal-weight)   & 73.9          & 74.4          & \textbf{93.9} & \textbf{98.2} & \textbf{95.8} & \textbf{98.9} & \textbf{99.5} & 77.9          & 88.9          \\
      EPO (equal-weight)  & \textbf{74.2} & \textbf{78.3} & 93.4          & \textbf{98.2} & 95.7          & 97.3          & 99.0          & \textbf{78.0} & \textbf{89.3} \\
      MGDA                & 64.8          & 63.7          & 87.7          & 92.3          & 90.8          & 98.4          & 99.0          & 74.7          & 83.9          \\
      \midrule
      Weight Averaging    & 65.3          & 63.3          & 71.4          & 73.6          & 64.2          & 52.8          & 87.5          & 50.1          & 66.0          \\
      Fisher Merging      & 68.6          & 69.2          & 70.7          & 66.4          & 72.9          & 51.1          & 87.9          & 59.9          & 68.3          \\
      RegMean             & 65.3          & 63.5          & 75.6          & 78.6          & 78.1          & 67.4          & 93.7          & 52.0          & 71.8          \\
      Task Arithmetic     & 55.3          & 54.9          & 66.7          & 77.4          & 80.2          & 69.7          & 97.3          & 50.1          & 69.0          \\
      Ties-Merging        & 65.0          & 64.3          & 74.7          & 76.8          & 81.3          & 69.4          & 96.5          & 54.3          & 72.8          \\
      AdaMerging          & 63.7          & 67.6          & 78.7          & 92.6          & 86.6          & \textbf{91.8} & \textbf{97.4} & 57.7          & 79.5          \\
      Pareto Merging      & 55.3          & 55.0          & 66.7          & 77.5          & 80.2          & 69.7          & 97.3          & 50.1          & 69.0          \\
      Ours-LS (MLP)       & 63.5          & 64.0          & 79.1          & 87.3          & 84.8          & 81.8          & 93.1          & 63.6          & 77.2          \\
      Ours-LS (Attn+MLP)  & 69.2          & 71.4          & \textbf{85.3} & \textbf{92.8} & 90.3          & 89.7          & 97.1          & \textbf{71.9} & \textbf{83.5} \\
      Ours-EPO (MLP)      & 66.2          & 65.9          & 77.5          & 82.7          & 81.6          & 79.4          & 88.2          & 64.3          & 75.7          \\
      Ours-EPO (Attn+MLP) & \textbf{70.0} & \textbf{72.2} & 85.0          & 90.7          & \textbf{90.9} & 87.2          & 94.0          & 70.6          & 82.6          \\
      \bottomrule
    \end{tabular}%
  }

  \vspace{0.8em}
  \fitwidth{%
    \begin{tabular}{@{}lccccccccc@{}}
      \toprule
      {Method}           & {SUN397}      & {Cars}        & {RESISC45}    & {EuroSAT}     & {SVHN}        & {GTSRB}       & {MNIST}       & {DTD}         & {Avg.}        \\
      \midrule
      Pretrained         & 68.2          & 77.9          & 71.3          & 61.3          & 58.4          & 50.6          & 76.4          & 55.4          & 64.9          \\
      Fine-tuned (STL)   & 82.3          & 92.4          & 97.4          & 99.9          & 98.1          & 99.2          & 99.7          & 84.1          & 94.1          \\
      \midrule
      LS (equal-weight)  & 80.8          & 90.6          & 96.3          & 96.3          & 97.6          & 99.1          & 99.6          & 84.4          & 93.5          \\
      \midrule
      Weight Averaging   & 72.1          & 81.6          & 82.6          & 91.4          & 78.2          & 70.6          & 97.0          & 62.8          & 79.5          \\
      Fisher Merging     & 69.2          & 88.6          & 87.5          & 93.5          & 80.6          & 74.8          & 93.3          & 70.0          & 82.2          \\
      RegMean            & 73.3          & 81.8          & 86.1          & \textbf{97.0} & 88.0          & 84.2          & 98.5          & 60.8          & 83.7          \\
      Task Arithmetic    & 74.1          & 82.1          & 86.7          & 92.6          & 87.9          & 86.8          & \textbf{98.9} & 65.6          & 84.4          \\
      Ties-Merging       & 75.0          & 84.5          & 88.0          & 94.3          & 85.7          & 82.1          & 98.7          & 67.7          & 84.5          \\
      AdaMerging         & 76.0          & 88.3          & 90.5          & 95.4          & 90.8          & \textbf{96.9} & \textbf{98.9} & 74.6          & 88.9          \\
      Pareto Merging     & 74.1          & 82.1          & 86.7          & 92.6          & 87.9          & 86.8          & \textbf{98.9} & 65.6          & 84.3          \\
      Ours-LS (MLP)      & 77.6          & 88.5          & 92.5          & 96.1          & 95.2          & 94.7          & 98.2          & 79.1          & 90.2          \\
      Ours-LS (Attn+MLP) & \textbf{79.3} & \textbf{90.0} & \textbf{94.1} & 96.8          & \textbf{96.6} & 96.4          & 98.7          & \textbf{81.3} & \textbf{91.6} \\
      \bottomrule
    \end{tabular}%
  }
\end{table*}

\begin{table*}[t]
  \centering
  \tablebodyfont
  \caption{
    Performance of CLIP-ViT-B/32 models on eight tasks with different preference vectors.
    Here the \textbf{largest} value and \textit{\underline{second largest}} value in each column are highlighted to show the trade-offs between tasks are learned successfully.
  }
  \label{table:8tasks}
  \fitwidth{%
  \begin{tabular}{@{}ccccccccc@{}}
    \toprule
    Preference Vector   & SUN397                    & Cars                      & RESISC45                  & EuroSAT                   & SVHN                      & GTSRB                     & MNIST                     & DTD                       \\ \midrule
    equal weights       & \textit{\underline{69.2}} & \textit{\underline{71.4}} & \textit{\underline{85.3}} & \textit{\underline{92.8}} & \textit{\underline{90.3}} & \textit{\underline{89.7}} & \textit{\underline{97.1}} & \textit{\underline{71.9}} \\
    $[0,0,0,0,0,0,0,1]$ & 59.9                      & 67.3                      & 71.2                      & 77.4                      & 72.1                      & 81.8                      & 90.3                      & \textbf{76.1}             \\
    $[0,0,0,0,0,0,1,0]$ & 67.2                      & 68.2                      & 84.1                      & 90.7                      & 84.9                      & 84.4                      & \textbf{99.3}             & 69.4                      \\
    $[0,0,0,0,0,1,0,0]$ & 65.1                      & 66.4                      & 79.1                      & 83.7                      & 74.7                      & \textbf{97.6}             & 90.2                      & 64.1                      \\
    $[0,0,0,0,1,0,0,0]$ & 65.4                      & 67.9                      & 79.3                      & 90.4                      & \textbf{95.5}             & 71.7                      & 85.6                      & 62.9                      \\
    $[0,0,0,1,0,0,0,0]$ & 65.7                      & 70.2                      & 71.1                      & \textbf{99.3}             & 81.6                      & 85.5                      & 94.9                      & 65.7                      \\
    $[0,0,1,0,0,0,0,0]$ & 58.7                      & 68.6                      & \textbf{93.8}             & 76.4                      & 82.4                      & 82.8                      & 92.9                      & 63.8                      \\
    $[0,1,0,0,0,0,0,0]$ & 56.7                      & \textbf{73.4}             & 73.6                      & 86.5                      & 82.1                      & 79.0                      & 89.4                      & 62.6                      \\
    $[1,0,0,0,0,0,0,0]$ & \textbf{69.5}             & 52.9                      & 62.4                      & 73.7                      & 77.4                      & 76.9                      & 95.2                      & 56.7                      \\
    \bottomrule
  \end{tabular}%
  }
\end{table*}

\textbf{Routing analysis}.
In order to gain deeper insight into the MoE modules' routing mechanism, we conduct an examination of the expert routing weights associated with various preference vectors, utilizing CLIP-ViT-B/32 (Ours-LS, MLP only).
For every unit preference vector, we calculate the routing weights at various depths for the MoE routers and display these weights in Figure~\ref{fig:vitb32_routing_weights}.
Notably, we have
(1) \textit{variability across depths}. The routing weights vary significantly across different depths of the model.
(2) \textit{Specificity to tasks}. Each preference vector, tied to a specific task, guides the routing towards relevant experts at higher depths. This shows the model's ability to adjust its weights according to the preference vector, confirming the intended behavior of MoE to specialize expert usage per task.
(3) \textit{Expert utilization}. Certain experts are utilized more often across various preference vectors, and this usage varies at different depths. For instance, the `Cars' expert is commonly used in early layers, suggesting its proficiency in extracting basic image features. Conversely, the `DTD' expert is often used in later layers, indicating its specialization in deeper contextual understanding.

\subsection{Natural Language Tasks}
\label{section:pareto_set_approximation_text_generation}

\begin{figure*}[t]
  \centering
  \includegraphics[width=0.90\linewidth]{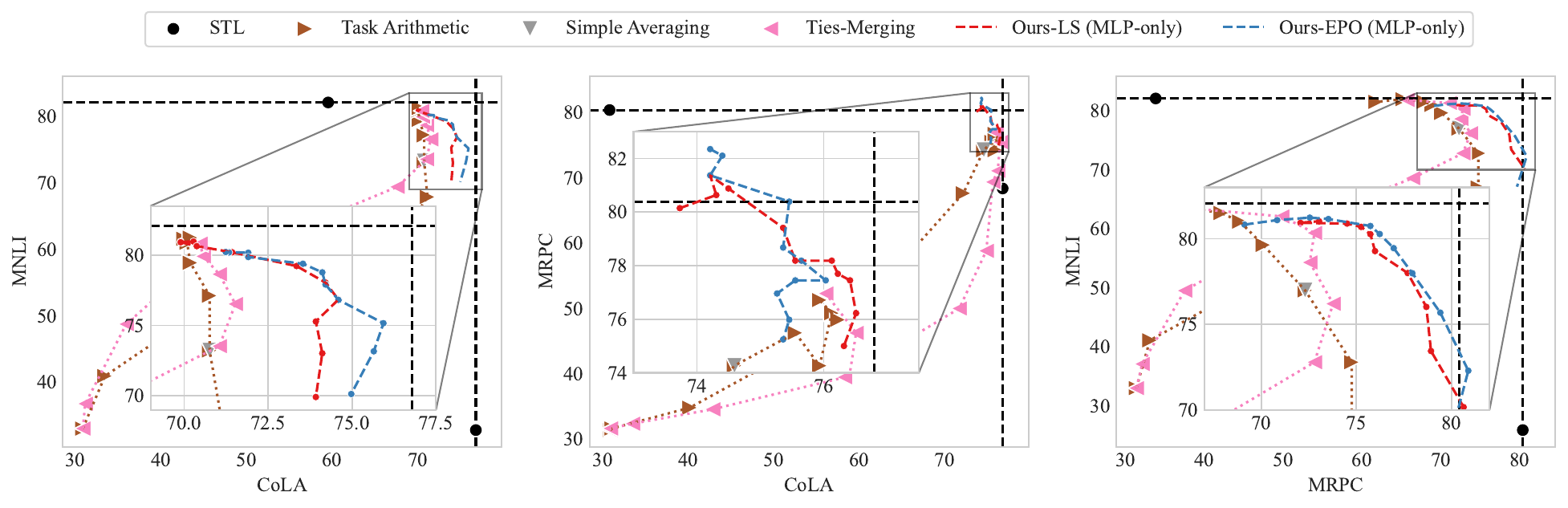}
  \caption{
    Visualization of results for natural language task pair experiments using GPT-2.
  }
  \label{fig:gpt2_pareto_moe_two_task}
\end{figure*}

In addition to image classification tasks, we also evaluate our method on natural language tasks using GPT-2 models.
(1) We first consider three task pairs from GLUE benchmark~\cite{wangGLUEMultiTaskBenchmark2018}, including CoLA-MNLI, CoLA-MRPC, and MRPC-MNLI.
We visualize the results using in Figure~\ref{fig:gpt2_pareto_moe_two_task}.
It shows a similar pattern to the image classification task pair experiments, where our approach successfully approximates the Pareto set, offering a good trade-off between the pair of tasks.
(2) We also scale up the number of objectives to seven and compare the results of an equal-weight preference vector with two model merging methods, Task Arithmetic and Ties-Merging, in Table~\ref{table:gpt2_seven_task}.

\begin{table*}[t]
  \tablebodyfont
  \centering
  \caption{Multi-task performance comparison on seven natural language tasks using GPT-2.}
  \label{table:gpt2_seven_task}
  \fitwidth{%
  \begin{tabular}{@{}lcccccccc@{}}
    \toprule
    {Method}         & {MRPC}        & {MNLI}        & {COLA}        & {SST-2}       & {QNLI}        & {QQP}         & {RTE}         & {Avg.}        \\
    \midrule
    Fine-tuned (STL) & 80.4          & 82.1          & 76.8          & 93.2          & 91.3          & 89.3          & 70.3          & 83.9          \\
    \midrule
    Task Arithmetic  & 67.9          & 68.1          & 68.8          & 85.1          & 70.1          & 80.9          & 47.3          & 69.4          \\
    Ties-Merging     & 68.4          & \textbf{71.4} & 68.4          & 81.8          & 69.6          & \textbf{82.4} & 47.7          & 70.0          \\
    Ours-LS (MLP)    & \textbf{77.0} & 68.8          & \textbf{72.5} & \textbf{85.2} & \textbf{75.6} & 81.1          & \textbf{62.8} & \textbf{74.7} \\
    \bottomrule
  \end{tabular}%
  }
\end{table*}

%% file: section/conclusion.tex
\section{Conclusion and Future Work}
\label{sec:conclusion}

In this work, we have introduced a novel approach that addresses the limitations of previous methods that either find a single Pareto optimal solution or require extensive computational resources to identify the entire Pareto set.
Our method, which leverages the Pareto-driven Weight-Ensembling Mixture of Experts (PWEMoE) structure, not only finds the Pareto set but also offers scalability to both the number of objectives and the size of the model.
By ensembling the weights of specialized single-task models, the MoE module can effectively capture the trade-offs between multiple objectives and provide a close approximation of the entire Pareto set of large neural networks.

Currently, our experiments are limited to Transformer-based models, such as ViT models and Transformer-based LLMs, further exploration can be conducted on other types of models, such as RNNs and CNNs, in future work.
While the method aims to approximate the Pareto front using a simple MLP-based MoE router, it may not always provide a complete or accurate representation of the Pareto set, further exploration can be conducted on more sophisticated router architectures.

%% file: section/appendix/appendix.tex


\section{Proof of Existence of an Error Bound}
\label{appendix:proof_of_error_bound}

Given a MOOP with $T$ objectives, the Pareto front is defined as $\mathcal{P} = \{\mathcal{L}(\phi) \mid \phi \in \Phi \text{ and } \nexists \phi' \in \Phi \text{ s.t. } \phi' \prec \phi \}$, where $\mathcal{L}(\phi) = (l_1(\phi), \ldots, l_T(\phi))$ is the vector of task-specific losses, and $\Phi=\mathbb{R}^{\vert\phi\vert}$ is the space of parameters $\phi$ for the upscaled module. The Pareto front estimated by the proposed MoE modules is defined as $\mathcal{P}_{\text{MoE}} = \{\mathcal{L}(\phi) \mid \phi \in \Phi_{\text{MoE}} \text{ and } \nexists \phi' \in \Phi_{\text{MoE}} \text{ s.t. } \phi' \prec \phi \}$.

Here we proof Theorem~\ref{theorem:existence_of_error_bound} as follows:

\begin{proof}
  Let $\phi^* \in \Phi_{MoE}$ be any point on the approximated Pareto front $\mathcal{P}_{MoE}$. Since for the MoE $\Phi_{\text{MoE}}$ is a subspace of the full parameter space $\Phi$, there exists a point $\phi \in \Phi$ that is closest to $\phi^*$ in Euclidean distance. That is,
  \begin{equation}\phi = \arg\min_{\phi' \in \Phi} \|\phi' - \phi^*\|_2.\end{equation}
  Since the loss functions $l_i$ are continuous, there exists an $\epsilon > 0$ such that
  \begin{equation}\|\phi^* - \phi\|_2 < \delta \implies \|l_i(\phi^*) - l_i(\phi)\|_2 < \epsilon \quad \forall i=1,\ldots,T.\end{equation}
  By the triangle inequality, this implies
  \begin{equation}
    \|\mathcal{L}(\phi^*) - \mathcal{L}(\phi)\|_2 = \sqrt{\sum_{i=1}^T (l_i(\phi^*) - l_i(\phi))^2} \leq \sqrt{T} \epsilon.
  \end{equation}
  Therefore, for any point $\mathcal{L}(\phi^*) \in \mathcal{P}_{\text{MoE}}$ on the Pareto front approximated by the MoE, there exists a point $\mathcal{L}(\phi) \in \mathcal{P}$ on the true Pareto front such that their Euclidean distance is bounded by $\sqrt{T} \epsilon$.
\end{proof}

\section{Task-Specific Model Acquisition}
\label{appendix:task_specific_model_acquistition}

\textbf{Image Classification.} For image classification tasks, we used the CLIP-ViT models~\cite{radfordLearningTransferableVisual2021} pre-trained on large-scale image-text datasets from OpenCLIP library~\cite{ilharco_gabriel_2021_5143773}.
Checkpoints of the fine-tuned models are the same as those used in the Task Arithmetic paper~\cite{ilharcoEditingModelsTask2023}, and are publicly available~\footnote{\url{https://github.com/mlfoundations/task_vectors}}.
In Tables~\ref{table:clip-vit-b-32_individuals} and~\ref{table:clip-vit-l-14_individuals}, we provide the individual performance of the pre-trained and fine-tuned CLIP-ViT-B/32 and CLIP-ViT-L/14 models on the eight image classification tasks.

\begin{table*}[t]
  \tablebodyfont
  \centering
  \caption{Individual performance of pre-trained and fine-tuned CLIP-ViT-B/32 models.}
  \label{table:clip-vit-b-32_individuals}
  \fitwidth{%
    \begin{tabular}{@{}lcccccccc@{}}
      \toprule
      \textbf{Model} & \textbf{SUN397} & \textbf{Cars} & \textbf{RESISC45} & \textbf{EuroSAT} & \textbf{SVHN} & \textbf{GTSRB} & \textbf{MNIST} & \textbf{DTD}  \\\midrule
      pre-trained    & 63.2            & 59.6          & 60.2              & 45.0             & 31.6          & 32.6           & 48.3           & 44.4          \\
      SUN397         & \textbf{75.3}   & 49.2          & 54.2              & 49.4             & 28.3          & 29.7           & 49.1           & 40.1          \\
      Cars           & 55.9            & \textbf{77.7} & 51.2              & 39.6             & 29.4          & 30.2           & 51.8           & 38.8          \\
      RESISC45       & 52.2            & 47.2          & \textbf{96.1}     & 56.3             & 24.2          & 22.5           & 49.6           & 34.7          \\
      EuroSAT        & 51.6            & 45.3          & 32.5              & \textbf{99.9}    & 19.3          & 26.1           & 37.9           & 35.9          \\
      SVHN           & 49.3            & 40.2          & 30.3              & 12.7             & \textbf{97.5} & 31.4           & 85.7           & 28.7          \\
      GTSRB          & 46.4            & 38.9          & 29.5              & 22.0             & 43.9          & \textbf{98.7}  & 39.5           & 28.5          \\
      MNIST          & 49.2            & 40.2          & 33.5              & 20.7             & 49.2          & 15.3           & \textbf{99.7}  & 27.4          \\
      DTD            & 50.4            & 49.4          & 41.9              & 33.9             & 28.9          & 22.8           & 47.8           & \textbf{79.4} \\\bottomrule
    \end{tabular}%
  }
\end{table*}

\begin{table*}[t]
  \tablebodyfont
  \centering
  \caption{Individual performance of pre-trained and fine-tuned CLIP-ViT-L/14 models.}
  \label{table:clip-vit-l-14_individuals}
  \fitwidth{%
    \begin{tabular}{@{}lcccccccc@{}}
      \toprule
      \textbf{Model} & \textbf{SUN397} & \textbf{Cars} & \textbf{RESISC45} & \textbf{EuroSAT} & \textbf{SVHN} & \textbf{GTSRB} & \textbf{MNIST} & \textbf{DTD}  \\\midrule
      pre-trained    & 68.2            & 77.9          & 71.3              & 61.3             & 58.4          & 50.6           & 76.4           & 55.4          \\
      SUN397         & \textbf{82.3}   & 71.2          & 64.7              & 54.6             & 52.5          & 46.9           & 75.1           & 51.9          \\
      Cars           & 67.2            & \textbf{92.4} & 68.4              & 56.4             & 57.8          & 48.4           & 73.7           & 55.6          \\
      RESISC45       & 66.3            & 71.5          & \textbf{97.4}     & 57.7             & 52.7          & 48.5           & 78.9           & 52.1          \\
      EuroSAT        & 65.7            & 70.8          & 46.9              & \textbf{99.9}    & 49.1          & 46.6           & 75.0           & 48.4          \\
      SVHN           & 67.6            & 70.8          & 64.4              & 37.0             & \textbf{98.1} & 47.2           & 91.1           & 51.9          \\
      GTSRB          & 66.5            & 73.4          & 64.8              & 34.5             & 61.6          & \textbf{99.2}  & 82.9           & 52.5          \\
      MNIST          & 68.5            & 73.0          & 65.5              & 43.4             & 66.5          & 44.1           & \textbf{99.7}  & 52.6          \\
      DTD            & 66.0            & 74.4          & 68.0              & 55.7             & 51.3          & 47.8           & 64.3           & \textbf{84.1} \\
      \bottomrule
    \end{tabular}%
  }
\end{table*}

\textbf{Natural Language Tasks.} For natural language processing tasks, we used the pre-trained GPT-2~\cite{radfordLanguageModelsAre2019} from the Hugging Face Transformers library~\cite{wolfHuggingFaceTransformersStateoftheart2020}. We fine-tuned the pre-trained GPT-2 model on the seven downstream tasks from the GLUE benchmark~\cite{wangGLUEMultiTaskBenchmark2018} for 3 epochs with a batch size of 8, the learning rate is set to $5 \times 10^{-5}$, and the Adam optimizer is used with weight decay of $10^{-9}$.
In Table~\ref{table:gpt-2_individuals}, we provide the individual performance of the fine-tuned GPT-2 models on the seven natural language processing tasks and bold the best performance for each task.

\begin{table*}[t]
  \tablebodyfont
  \centering
  \caption{Individual performance of pre-trained and fine-tuned GPT-2 models.}
  \label{table:gpt-2_individuals}
  \fitwidth{%
    \begin{tabular}{@{}lcccccccc@{}}
      \toprule
      \textbf{Model} & \textbf{MRPC} & \textbf{MNLI} & \textbf{COLA} & \textbf{SST-2} & \textbf{QNLI} & \textbf{QQP}  & \textbf{RTE}  & \textbf{Avg.} \\
      \midrule
      MRPC           & \textbf{80.4} & 25.9          & 30.8          & 49.1           & 47.1          & 65.9          & 49.1          & 49.8          \\
      MNLI           & 33.8          & \textbf{82.1} & 59.5          & 40.5           & 46.5          & 24.9          & 57.4          & 49.2          \\
      COLA           & 68.4          & 32.8          & \textbf{76.8} & 51.0           & 50.4          & 39.2          & 48.0          & 52.4          \\
      SST-2          & 40.2          & 32.9          & 51.8          & \textbf{91.2}  & 49.8          & 56.8          & 44.4          & 52.4          \\
      QNLI           & 30.6          & 38.9          & 58.7          & 47.0           & \textbf{88.3} & 39.9          & 48.7          & 50.3          \\
      QQP            & 62.3          & 25.7          & 31.4          & 49.1           & 45.0          & \textbf{89.6} & 49.1          & 50.3          \\
      RTE            & 37.5          & 47.7          & 52.8          & 54.9           & 53.5          & 33.7          & \textbf{65.3} & 49.3          \\
      \bottomrule
    \end{tabular}%
  }
\end{table*}

\section{Model Merging Baselines}
\label{appendix:model_merging_baselines}

Here we provide a brief overview of the model merging baselines used in our experiments.
We compare our approach with several established model merging techniques that aggregate knowledge from multiple models to enhance performance across different tasks. Below are the details of each baseline method:

\begin{enumerate}
  \item \textbf{Simple Averaging}: This method merges the paramters of multiple models by taking the average of their weights.
  \item \textbf{Fisher Merging}~\cite{matenaMergingModelsFisherWeighted2022}: This approach combines models by taking the weighted average of their parameters based on the Fisher information matrix.
        For each fine-tuned model, compute the Fisher information matrix, which captures the importance of each model.
  \item \textbf{RegMean}~\cite{jinDatalessKnowledgeFusion2023}: RegMean is a model merging technique for merging linear layers. RegMean operates by pre-computing the inner production matrices of training data for each individual model. When it comes to merging these models, the method retrieves the weights and inner product matrices from each model and computes new weights based on equation:
        \begin{equation}
          W_{\text{merged}} = \left(\sum_i X_i^\top X_i\right)^{-1}\sum_i \left(X_i^\top X_i W_i\right).
        \end{equation}
  \item \textbf{Task Arithmetic}~\cite{ilharcoEditingModelsTask2023}: For each model and task pair, we generate a task vector by taking the element-wise subtraction of the task-specific model and the pre-trained model within the parameter space. Subsequently, the task vectors from all models are summed up into a single vector, multiplied by a scalar factor, and added to the pre-trained model to obtain the merged model. Mathematically, the merging process can be represented as:
        \begin{equation}
          W_{\text{merged}} = W_{\text{pre-trained}} + \lambda \sum_i \left(W_{i} - W_{\text{pre-trained}}\right),
        \end{equation}
        where $\lambda$ is the scaling coefficient and $W$ denotes the weights of the model.
  \item \textbf{Ties-Merging}~\cite{yadavResolvingInterferenceWhen2023}: Ties-Merging is a model merging technique like Task Arithmetic, but it uses a different strategy to merge the task vectors. Instead of summing the task vectors, Ties-Merging takes a more nuanced approach.
        In Ties-Merging, the process starts with ``Trimming'', discarding minor updates to the fine-tuned model's parameters to reduce noise.
        Then, ``Elect Sign'' determines the sign of each parameter update based on its impact on task performance, resolving sign disagreements across the models either by majority vote or using an oracle sign vector.
        Lastly, ``Merge'' combines the parameter updates respecting the resolved signs, aiming to maintain the specialized learning from each task and create a multi-task model without additional fine-tuning.
\end{enumerate}

\begin{figure*}[htbp]
  \centering
  \includegraphics[width=0.82\linewidth]{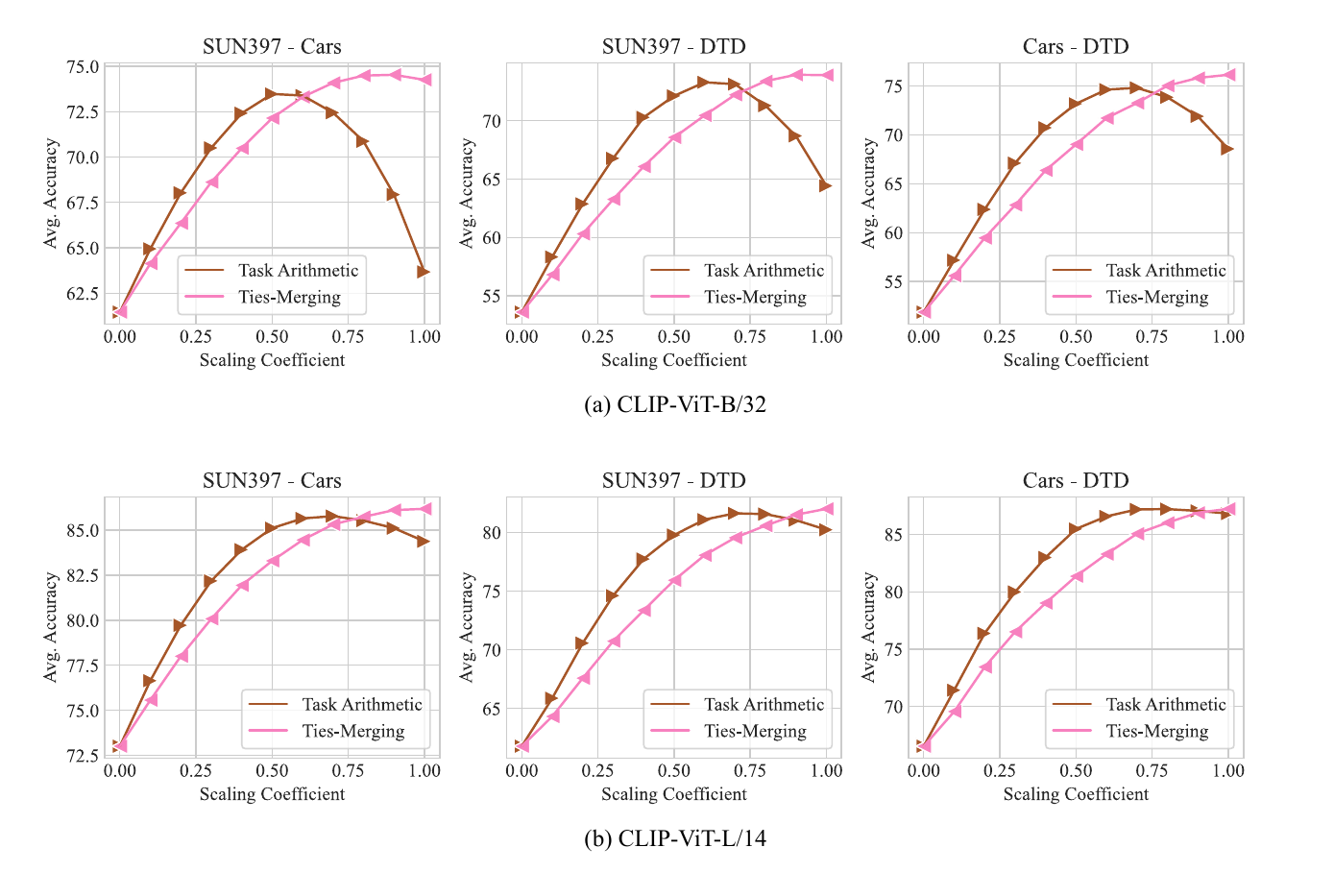}
  \caption{Merging two models for image classification tasks using Task Arithmetic and Ties-Merging.
    The horizontal axis represents scaling coefficients, and the vertical axis indicates the average accuracy of the merged model on the two tasks. It can be observed that for the two tasks scenario, Task Arithmetic roughly achieves the best performance around $\lambda=0.6$, while Ties-Merging performs best at $\lambda=1$.}
  \label{fig:clip_two_tasks_task_arithmetic_and_ties_merging}
\end{figure*}

These baseline methods serve as important benchmarks to assess the performance gains obtained by our proposed approaches.
In particular, we inverstigate the performance of Task Arithmetic and Ties-Merging on the two-tasks scenario for CLIP-ViT-B/32 and CLIP-ViT-L/14 models in Figure~\ref{fig:clip_two_tasks_task_arithmetic_and_ties_merging}.
Because for these two methods, the performance of the merged model is highly dependent on the scaling coefficient $\lambda$, we plot the average accuracy of the merged model on the two tasks against the scaling coefficient $\lambda$.
It can be observed that for the two-tasks scenario, Task Arithmetic roughly achieves the best performance around $\lambda=0.6$, while Ties-Merging performs best at $\lambda=1$.
In our experiments, we report the best performance of each method on the validation set.

\section{Pareto Optimal Learning Baselines}
\label{appendix:pareto_optimal_learning_baselines}

\textbf{Pareto set learning methods.}
There are many methods for learning the Pareto set of a multi-objective optimization problem, as introduced in Section~\ref{sec:related_work}.
Some methods are theoretically solid and elegant.
However, most of these methods demand significant computational resources and do not scale well to larger models.
Typically, these methods are limited to benchmarking in multi-objective optimization problems with 2-3 closed-form objective functions.
Therefore, we do not compare with these methods in our experiments.
In this work, we focus on the Pareto optimal learning methods that can be applied to large-scale neural networks, such as linear scalarization, EPO serach and MGDA.

\begin{enumerate}
  \item \textbf{Linear Scalarization.} This method involves combining the multiple objectives into a single objective function by assigning a weight to each objective. The weights are usually determined based on the importance of each objective. The combined objective function is then optimized using standard optimization techniques. However, this method can only find solutions that lie on the convex hull of the Pareto front. The single objective function is defined as follows:
        \begin{equation}
          l_{LS} = \sum_{i=1}^T r_i l_i(x),
        \end{equation}
        where $l_i(x)$ is the $i$-th objective function, $r_i$ is the weight assigned to the $i$-th objective, and $T$ is the number of objectives.
  \item \textbf{EPO Search}~\cite{mahapatraMultiTaskLearningUser2020}.
        The Exact Pareto Optimal (EPO) search is also a method that aims to find a single Pareto optimal solution.
        Unlike linear scalarization, EPO search can find solutions that are exactly Pareto optimal with respect to given preference vector.
        Where the non-uniformity of the objective values is taken into account, which is defined as follows:
        \begin{equation}
          \text{Non-uniformity} = \text{KL}\left(\hat{l} \;\middle\vert\; \frac{1}{T} \right).
        \end{equation}
        Where $\hat{l}$ is the weighted normalization of the objective values, and $T$ is the number of objectives.
        \begin{equation}\hat{l}_i = \frac{r_i l_i}{\sum_{j=1}^T r_j l_j}\end{equation}
  \item \textbf{MGDA.}
        The Multiple Gradient Descent Algorithm (MGDA) is a method that uses the gradients of the objective functions to find a direction that improves all objectives simultaneously.
        In our experiments, we compare the performance of our proposed methods with these Pareto optimal learning baselines. The results show that our methods can achieve comparable or better performance, demonstrating their effectiveness in MOOPs.
\end{enumerate}

\section{Device Specifications}
\label{appendix:device_specifications}

All experiments were carried out on a server with 4 NVIDIA 4090 GPUs, each with 24GB of memory, using PyTorch~\cite{paszkePyTorchImperativeStyle2019a} and Hugging Face Transformers~\cite{wolfHuggingFaceTransformersStateoftheart2020} libraries.

\textbf{Computational burden.}
The computational burden of different methods can be influenced by several factors, including the number of trainable parameters, the number of objectives, and the size of the model.
In our experiments, we compare the computational burden of our proposed methods with the Pareto optimal learning baselines, including linear scalarization (LS), EPO search, and MGDA.
In Table~\ref{table:training_details_image_classification}, we summarize the training details for the Pareto set approximation experiments with image classification tasks.

\begin{table*}[htbp]
  \tablebodyfont
  \caption{
    Summarization of training details for Pareto set approximation experiments with image classification tasks.
    We use Adam optimizer for all experiments.
    The training time is measured on NVIDIA 4090 GPUs with 24GB memory.
    We utilize a CPU implementation of the EPO method, leading to longer training time. This can be further optimized.
  }
  \label{table:training_details_image_classification}
  \fitwidth{%
    \begin{tabular}{@{}lccccl@{}}
      \toprule
      Method         & \#Tasks & {\#Trainable / \#Total}     & {Batch Size (each)} & GPU usage             & Wall Time          \\
      \midrule
      \multicolumn{6}{c}{\textit{CLIP-ViT-B/32, 113.45M params.}}                                                               \\
      LS             & 2       & 113.45M / 113.45M (100\%)   & 16                  & 3.4GB                 & $\approx$ 4 mins   \\
      EPO            & 2       & 113.45M / 113.45M (100\%)   & 16                  & 6.1GB                 & $\approx$ 8 mins   \\
      Ours-LS        & 2       & 264 / 226.79M  (0.0001\%)   & 16                  & 2.8GB                 & $\approx$ 2-3 mins \\
      Ours-LS (All)  & 2       & 1.43K / 338.88M (0.0004\%)  & 16                  & 3.9GB                 & $\approx$ 3 mins   \\
      Ours-EPO       & 2       & 264 / 226.79M  (0.0001\%)   & 16                  & 3.1GB                 & $\approx$ 3-4 mins \\
      Ours-EPO (All) & 2       & 1.43K / 338.88M (0.0004\%)  & 16                  & 3.9GB                 & $\approx$ 5-6 mins \\
      Ours-LS        & 3       & 540 / 283.46M  (0.0002\%)   & 16                  & 3.6GB                 & $\approx$ 3 mins   \\
      Ours-EPO       & 3       & 540 / 283.46M  (0.0002\%)   & 16                  & 3.9GB                 & $\approx$ 5 mins   \\
      LS             & 8       & 113.45M / 113.45M (100\%)   & 16                  & 7.4GB                 & $\approx$ 13 mins  \\
      Ours-LS        & 8       & 3.36K / 566.81M  (0.0006\%) & 16                  & 7.4GB                 & $\approx$ 10 mins  \\
      Ours-LS (All)  & 8       & 18.20K / 1.02B  (0.0018\%)  & 16                  & 10.0GB                & $\approx$ 8 mins   \\
      Ours-EPO       & 8       & 3.36K / 566.81M (0.0006\%)  & 16                  & 7.8GB                 & $\approx$ 30 mins  \\
      Ours-EPO (All) & 8       & 18.20K / 1.02B (0.0018\%)   & 16                  & 10.5GB                & $\approx$ 85mins   \\
      \multicolumn{6}{c}{\textit{CLIP-ViT-L/14, 342.56M params.}}                                                               \\
      EPO            & 2       & 342.56M / 342.56M  (100\%)  & 16                  & 4$\times$20.5GB (DDP) &                    \\
      MGDA           & 2       & 342.56M / 342.56M  (100\%)  & 16                  & 2$\times$20.5GB (DDP) & $\approx$ 80 mins  \\
      Ours-LS        & 2       & 528 / 745.46M  (0.0001\%)   & 16                  & 19.8GB                & $\approx$ 16 mins  \\
      Ours-EPO       & 2       & 528 / 745.46M  (0.0001\%)   & 16                  & 20.8GB                & $\approx$ 37 mins  \\
      Ours-LS        & 8       & 6.72K / 1.95B  (0.0003\%)   & 12                  & 4$\times$21.5GB (DDP) & $\approx$ 15 mins  \\
      Ours-EPO       & 8       & 6.72K / 1.95B  (0.0003\%)   & 12                  & 4$\times$22.1GB (DDP) & $\approx$ 85 mins  \\
      \bottomrule
    \end{tabular}%
  }
\end{table*}

Impirically, it is observed that the computational burden can be aranged as follows: Ours-LS(MLP) $<$ Ours-EPO(MLP) $<$ LS $<$ Ours-LS(Attn+MLP) $<$ Ours-EPO(Attn+MLP) $<$ EPO $<$ MGDA.
The EPO method, on the other hand, tends to have a higher computational burden because we use a CPU implementation of the EPO method, leading to longer training time, which can be further optimized.

\textbf{Efficiency of the proposed methods.}
The low computational burden of the proposed methods can be attributed to the efficient nature of the weight-ensembling mixture of experts (MoE) structure. This structure allows for tuning only a small number of parameters to capture the trade-offs between multiple objectives. By ensembling the weights of specialized single-task models, the MoE module can provide a close approximation of the entire Pareto set of large neural networks, thus reducing the computational demand and memory consumption compared to other methods. This efficiency makes the approach scalable to both the number of objectives and the size of the model, making it a practical solution for multi-objective optimization problems in deep learning.

\section{Routing Analysis}
\label{appendix:routing_analysis}

To better understand the routing mechanism in the weight-ensembling Mixture of Experts (MoE) modules, we analyze the expert routing weights associated with different preference vectors using CLIP-ViT-B/32 (Ours-LS, MLP only). For each unit preference vector, we compute the routing weights at various depths for the MoE routers and present these weights in Figure~\ref{fig:vitb32_routing_weights}. We observe the following key points:

\begin{figure*}[htb]
  \centering
  \includegraphics[width=0.72\linewidth]{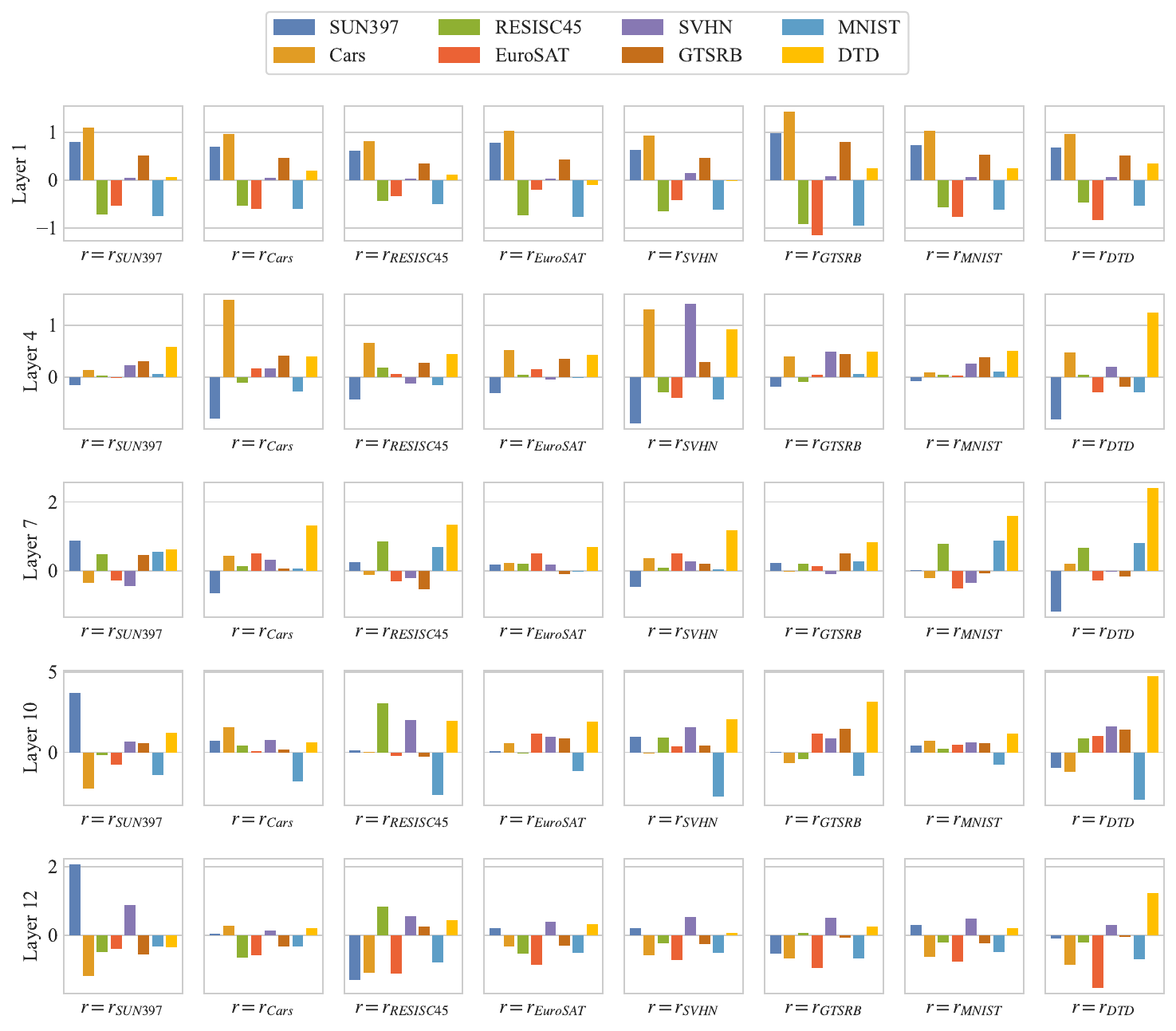}
  \caption{
    Visualization of routing weights for different perference vectors using CLIP-ViT-B/32. The preference vector for a specific task is denoted as $r_{task}$, with a value of one, and zeros for others.
  }
  \label{fig:vitb32_routing_weights}
\end{figure*}

\begin{itemize}
  \item  \textbf{Variability across depths}:
        The routing weights exhibit significant variation across different depths of the model.
        According to the impirical findings of previous research~\cite{zeilerVisualizingUnderstandingConvolutional2013},
        at various depths, the model processes information at different levels of abstraction. In early layers, the model may focus on low-level features like edges and textures, while in later layers, it may capture more complex and high-level concepts. This variation in the level of abstraction can lead to differences in the routing weights across depths.
  \item  \textbf{Specificity to tasks}:
        Each preference vector, associated with a specific task, directs the routing towards relevant experts at higher depths.
        This demonstrates the model's capability to adapt its weights based on the preference vector, validating the intended behavior of MoE to specialize expert usage for each task.
  \item \textbf{Expert utilization}:
        Some experts are more frequently utilized across different preference vectors, and this usage varies at different depths. For example, the `Cars' expert is frequently employed in early layers, indicating its effectiveness in extracting basic image features. In contrast, the `DTD' expert is often used in later layers, suggesting its specialization in deeper contextual understanding.
\end{itemize}